\documentclass{article}

\usepackage{microtype}
\usepackage{graphicx}
\usepackage{subfigure}
\usepackage{booktabs} % for professional tables
\usepackage{multicol}
\usepackage{multirow}
\usepackage{subfigure}
\usepackage{hyperref}

\usepackage[accepted]{icml2024}

\usepackage{amsmath}
\usepackage{amssymb}
\usepackage{mathtools}
\usepackage{amsthm}
\usepackage{comment}
\usepackage[capitalize,noabbrev]{cleveref}

\usepackage{siunitx}
\usepackage{tablefootnote}

\newcommand{\wrt}{\emph{w.r.t.}}
\newcommand{\eg}{\emph{e.g.}}
\newcommand{\ie}{\emph{i.e.}}
\newcommand{\etc}{\emph{etc}}
\theoremstyle{plain}

\theoremstyle{definition}

\theoremstyle{remark}

\usepackage[textsize=tiny]{todonotes}

\icmltitlerunning{Rethinking Optimization and Architecture for Tiny Language Models}

\begin{document}

\twocolumn[
\icmltitle{PanGu-$\pi$ Pro: Rethinking Optimization and Architecture  for \\ Tiny Language Models}




\begin{icmlauthorlist}
\icmlauthor{Yehui Tang}{comp1}
\icmlauthor{Kai Han}{comp1}
\icmlauthor{Fangcheng Liu}{comp1}
\icmlauthor{Yunsheng Ni}{comp1}
\icmlauthor{Yuchuan Tian}{comp2}
\icmlauthor{Zheyuan Bai}{comp1}
\icmlauthor{Yi-Qi Hu}{comp3}
\icmlauthor{Sichao Liu}{comp3}
\icmlauthor{Shangling Jui}{comp4}
\icmlauthor{Yunhe Wang}{comp1}
\end{icmlauthorlist}

\icmlaffiliation{comp1}{Huawei Noah’s Ark Lab}
\icmlaffiliation{comp2}{Peking University}
\icmlaffiliation{comp3}{Consumer Business Group, Huawei}
\icmlaffiliation{comp4}{Huawei Kirin Solution}
\icmlcorrespondingauthor{Yunhe Wang}{yunhe.wang@huawei.com}

\icmlkeywords{Machine Learning, ICML}

\vskip 0.3in
]



\printAffiliationsAndNotice{}  % leave blank if no need to mention equal contribution

\begin{abstract}
The power of large language models (LLMs) has been demonstrated through numerous data and computing resources. However, the application of language models on mobile devices is facing huge challenge on the computation and memory costs, that is, tiny language models with high performance are urgently required. Limited by the highly complex training process, there are many details for optimizing language models that are seldom studied carefully. In this study, based on  a tiny language model with 1B parameters, we carefully design a series of empirical study to analyze the effect of each component. Three perspectives are mainly discussed, \ie, neural architecture, parameter initialization, and optimization strategy. Several design formulas are empirically proved especially effective for tiny language models,  including tokenizer compression, architecture tweaking, parameter inheritance and multiple-round training. Then we  train  PanGu-$\pi$-1B Pro and PanGu-$\pi$-1.5B Pro on 1.6T multilingual corpora, following the established formulas.  Experimental results demonstrate the improved optimization and architecture yield  a notable average improvement of 8.87 on benchmark evaluation sets for PanGu-$\pi$-1B Pro. Besides, PanGu-$\pi$-1.5B Pro surpasses a range of SOTA models with larger model sizes, validating its superior performance. The code is available\footnote{\url{https://github.com/YuchuanTian/RethinkTinyLM}}.
\end{abstract}

\section{Introduction}
\label{sec:intro}

Large language models (LLMs), trained on extensive corpora, have demonstrated impressive performance across diverse natural language tasks. The release of ChatGPT, with its robust generalization capabilities, has captured global attention and holds the potential to revolutionize the interaction between humans and computers.

\begin{figure}[t]

	\begin{center}
		\centerline{\includegraphics[width=0.99\columnwidth]{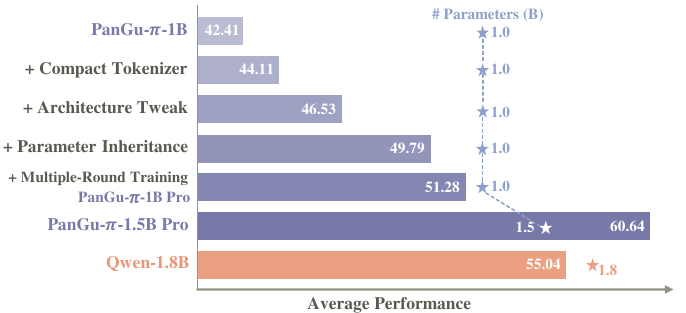}}
		%\vspace{-10pt}
		\caption{PanGu-$\pi$ Pro with improved architecture and optimization methods. PanGu-$\pi$-1B~\cite{wang2023PanGu} directly use the developing strategies of LLMs while PanGu-$\pi$-1B Pro achieves an average performance improvement of 8.87 with our methodology. It is worth mentioning that PanGu-$\pi$-1.5B Pro outperforms Qwen-1.8B~\cite{qwen} with 16.67\% fewer parameters.}
		\label{img:overall}
	\end{center}
	%\vspace{-25pt}
\end{figure}

In addition to the GPT-series models~\cite{radford2018improving, brown2020language,achiam2023gpt}, various large language models have emerged. 
PaLM~\cite{chowdhery2023palm}  trains a model with an impressive 540B parameters across 6144 TPU v4 chips. LLaMA~\cite{touvron2023llama}  releases a series of foundational language models, ranging from 7B to 70B parameters. Both the model architecture and trained weights are open-source, fostering collaboration within the AI community. Most of the following large models leverage similar architectures and training methodologies. For instance, Baichuan teams~\cite{yang2023baichuan} train 7B and 13B parameter models on a 2.6T token dataset encompassing both Chinese and English corpora. Qwen~\cite{qwen}, Yi~\cite{Yi}, and Skywork~\cite{wei2023skywork}  pursue similar paths, training models with 2.4T, 3T, and 3.2T tokens, respectively. Primarily attributed to the increasing accumulation of  cleaned data, the performance of LLMs improves rapidly.

While numerous studies have successfully trained various high-performance language models~\cite{ren2023PanGu, zeng2022glm}, the methodologies employed in training such models remain insufficiently analyzed. On one hand, a substantial body of work concentrates on collecting and cleaning  data, with less emphasis on researching effective training strategies. On the other hand, the training of large models demands an exceedingly high computational resource investment, making it impractical to explore a wide range of optimization strategies. As a result, recent works often adopt similar training recipes when constructing LLMs~\cite{touvron2023llama,Yi,qwen,wei2023skywork}.

Moreover, the implementation of these large models demands prohibitively high memory and computational resources, constraining their practical applicability in various scenarios. For example, the GPT-3 with 175B parameters necessitates approximately 700GB of memory when stored with FP32 datatype. Although the 7B parameter models are relatively more efficient, their resource requirements still render them impractical for deployment on edge devices, such as mobile phones.

In this paper, we  systematically rethink the methodology for constructing a tiny language model, including neural architecture, parameter initialization, and optimization strategy:
\begin{itemize} 
	\item Neural architecture: Adopting the tokenizer directly from larger models introduces redundant parameters, resulting in increased computational overhead. Streamlining the tokenizer by removing low-frequency vocabularies enhances the model's representational efficiency.  Moreover, we observe that the configuration of the model's architecture (depth, width, and expanding rate in FFN) has a significant impact on the final performance. Depth is the primary factor for tiny language models, and deeper models usually achieve high performance at the expense of lower inference speed.
	\item Parameter initialization: Inheriting parameters from the large model proves effective in boosting performance and expediting convergence.  The identification of crucial parameters is imperative in this context.  We have observed that layers situated near the beginning and end of the model often carry more significance than the intermediate layers. Furthermore, within each layer, the adoption of data-driven learnable criteria has demonstrated greater efficacy compared to heuristic methods.  
	\item Model optimization: In comparison to larger models, tiny models face more severe data forgetting issues, and multiple-round training proves beneficial for memory enhancement. We propose a straightforward sample selection strategy to mitigate the training cost associated with multiple-round training. Besides, we also delve into the relationship between batch size and learning rate specifically for tiny models.
\end{itemize}

Drawing from the aforementioned insights, we develop PanGu-$\pi$-1B Pro and PanGu-$\pi$-1.5B Pro with enhanced architecture and optimization methods.  From the developing strategies of LLMs, we gradually add four core components to improve performance (see Figure~\ref{img:overall}). 
The models are evaluated on various benchmarks including examination, knowledge, reasoning, and understanding, where our models achieve SOTA performance when compared with models of similar sizes. For instance, with 16.67\% fewer parameters, PanGu-$\pi$-1.5B Pro achieves an average score of 60.64, outperforming Qwen-1.8B which achieves a score of 55.04.

\section{Neural Architecture}

In this section, we investigate the architecture design of tiny language models. The experiments are conducted on 50B tokens randomly sampled from the pre-trained dataset, with equal proportions of Chinese and English corpus. The baseline is a 1B parameter model with LLaMA-like architecture unless specified. The models constructed with different strategies are compared on ARC Easy~\cite{clark2018think}, HellaSwag~\cite{zellers2019hellaswag} and C3~\cite{sun2019investigating}.

\subsection{Compact Tokenizer}
\label{Compact Tokenizer}

The tokenizer serves to map original natural language into tokens suitable for processing by large language models, with each token representing a word, subword, character, or symbol.
A multilingual tokenizer typically has a large vocabulary to cover various corpora. 
However, in the context of a tiny language model, an overly large vocabulary can significantly occupy a substantial portion of the model's parameters.   For instance, Qwen-7B~\cite{qwen}, Baichuan2-7B~\cite{yang2023baichuan}, and PanGu-$\pi$-7B~\cite{wang2023PanGu} have vocabulary sizes of 151936, 125696,  100883, respectively.  The parameters of their heads and embedding layers account for 16.12\% ,13.72\%, 10.91\% of the overall parameters.   While the PanGu-$\pi$-1B model with 12 layers and a width of 2048, using the same tokenizer, sees the head and embedding layers' parameters comprising a substantial 36.8\% of the total (Figure~\ref{img:tokenizer-ratio}). This distribution leads to a significant allocation of parameters to vocabulary representation rather than the main body, potentially limiting the model's overall representation capacity. Therefore, compressing the tokenizer becomes essential for a tiny language model to reduce its parameter proportion. 

\begin{figure}[t]
	%\vskip 0.2in
	\begin{center}
		\centerline{\includegraphics[width=0.8\columnwidth]{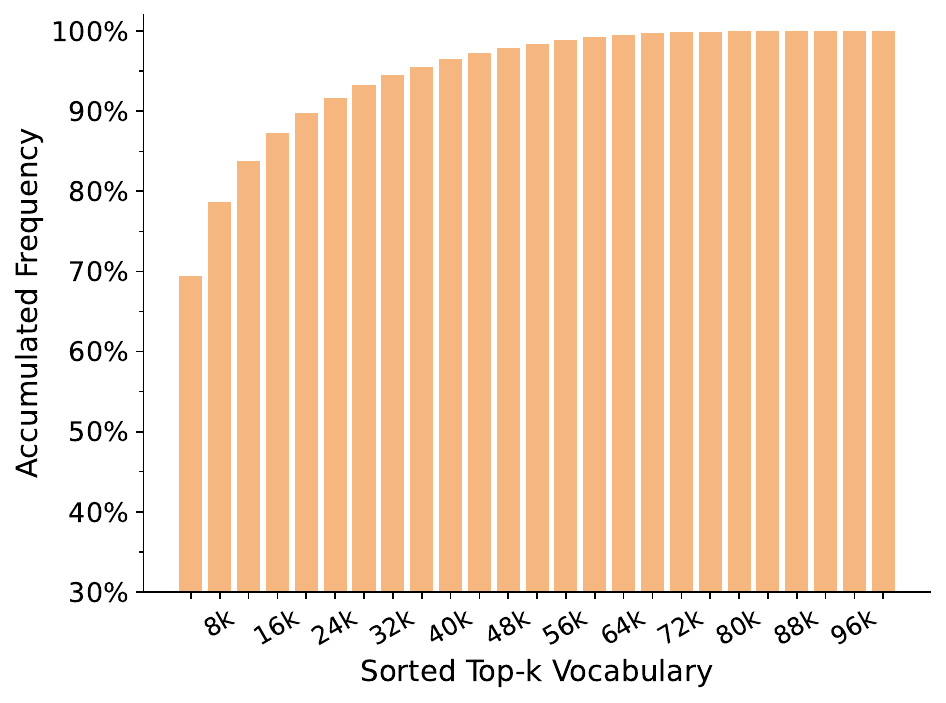}}
	%	\vspace{-15pt}
		\caption{Accumulative frequency of the top-k vocabularies, where 97.86\% data can be represented by a small 48k tokenizer.}
		\label{img:tokenizer-frequency}
	\end{center}
	%\vspace{-20pt}
\end{figure}

\begin{figure}[t]
	%\vskip 0.2in
	\begin{center}
		\centerline{\includegraphics[width=0.95\columnwidth]{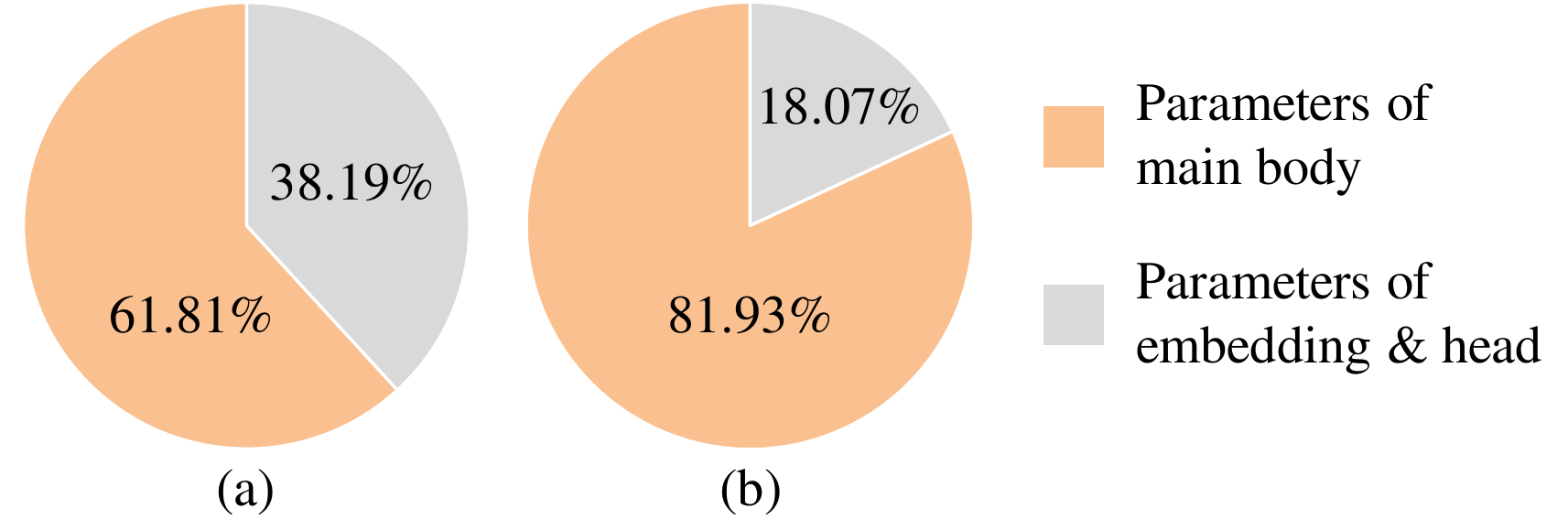}}
	%	\vspace{-10pt}
		\caption{The parameter proportions of model's main body and tokenizer. (a) The large tokenizer inherited from large multilingual models~\cite{wang2023PanGu}. (b) Compact tokenizer by removing low-frequency vocabularies.}
		\label{img:tokenizer-ratio}
	\end{center}
%	\vspace{-25pt}
\end{figure}

Actually, we discover that substantial redundancy exists in the tokenizer. By initializing the tokenizers with the 100k vocabularies inherited from the PanGu-$\pi$ model, we conducted a frequency analysis across a vast corpus comprising approximately 1.6T tokens.  As depicted in Figure~\ref{img:tokenizer-frequency}, it is evident that tokens exhibit a long-tail effect, where the top 48k vocabularies accounting for 97.86\% of all the training corpus. We conduct experiments with six vocabulary sizes $\{\text{8k}, \text{16k}, \text{32k}, \text{48k}, \text{72k}, \text{100k}\}$, which account for 78.68\%, 87.24\%, 94.49\%, 97.86\%, 99.84\% and 100\% accumulated frequency respectively. Over 50\% vocabularies may be redundant as they cater to less than 3\% of the corpus.

We advocate for the removal of low-frequency vocabularies to reduce their parameters. Table~\ref{tab:tokenizer_pref}  illustrates the performance variations concerning tokenizer size\footnote{All model's sizes are controlled to 1B by adjusting depth.}. 
The embedding and head layers constitute 18.07\% of the 1B model's parameters when using a vocabulary of 48k, showcasing the best average performance followed by the model with a vocabulary of 32k. It is noteworthy that employing an excessively small vocabulary can result in performance degradation. For instance, with an 8k tokenizer covering less than 70\% of the corpus, the model exhibits subpar performance on C3 and ARC-E datasets. The tokenizer with a size of 48k also exhibits a similar compression rate to that of the original 100k size tokenizer, which is evaluated across the entire training corpus.  Therefore, we recommend using a compact tokenizer covering over 90\% of the corpus, while ensuring the parameter proportion of embedding and head layers remains below 20\%.

\begin{table}[t]
	\centering
	\caption{Performance varies \wrt~tokenizer size. PEHF stands for the proportion of embedding and head layers over the whole model.}
	\label{tab:tokenizer_pref}%
	\resizebox{0.99\columnwidth}{!}{
		\begin{tabular}{cc|ccc|c}
			\toprule
			Tokenizer & PEHL (\%) & ARC-E & HellaSwag & C3 & Avg.   \\
			\midrule
			8k & 2.97 & 31.39 & 40.19 & 42.25 & 37.94 \\ 
			16k & 6.01 & 30.34 & 40.10 & 45.64 & 38.69 \\ 
			32k & 11.79 & 34.45 & 40.23 & 46.77 & 40.48 \\ 
			48k & 18.07 & 34.39 & \textbf{41.48}  & \textbf{47.70} & \textbf{41.19} \\ 
			72k & 26.88 & 34.39 & 39.21 & 46.58 & 40.06 \\ 
			100k & 38.19 & \textbf{34.98} & 39.11 & 47.10 & 40.40 \\ 
			\bottomrule
		\end{tabular}%
	}
%	\vspace{-10pt}
\end{table}%

\subsection{Architecture Tweak}
\label{sec:arch}

\begin{table}[t]
	\centering
%	\vspace{-5pt}
	\caption{Varying the depth and width of a 1B-size model with fixed vocabulary size  and expanding rate. The speed is measured by tokens per second.}
	\label{tab:with-depth-ratio}%
	\resizebox{0.99\columnwidth}{!}{
		\begin{tabular}{ccc|ccc|c}
			\toprule
			Depth & Width  & Speed  & ARC-E & HellaSwag  & C3  & Avg. \\
			\midrule
			40    & 1280     &  12.81  & \textbf{37.01} &  41.00 & \textbf{48.05}   & \textbf{42.02} \\
			30    & 1536     &  17.71  & 36.16 & 40.32  & 47.84   & 41.44\\
			20    & 1792     &  29.49  & 34.39 & \textbf{41.48}  & 47.70 & 41.19 \\ 
			15    & 2048     &   36.79 & 32.45 & 40.22  & 40.05    & 37.57 \\
			9     & 2560      &  \textbf{57.53}    & 32.63 & 31.06 & 42.68   &  35.46\\
			\bottomrule
		\end{tabular}%
	}
%	\vspace{-15pt}
\end{table}%

\begin{table}[t]
	\centering
%	\vspace{-5pt}
	\caption{Varying the expanding rate and width of a 1B-size model with fixed vocabulary size and depth.}
	\label{tab:expanding rate}%
	\resizebox{0.99\columnwidth}{!}{
		\begin{tabular}{ccc|ccc|c}
			\toprule
			EP Rate & Width & Speed & ARC-E & HellaSwag   & C3  & Avg.  \\
			\midrule
			1.00     & 2304  &     28.39      & 31.75 & 38.71 & 42.68   & 37.71 \\
			2.00     & 2048  &      28.68      & 33.33 & 41.34   & \textbf{48.55} &  41.07 \\
			2.77  & 1792       &  28.40 & 34.39 & \textbf{41.48}  & 47.70 & \textbf{41.19} \\ 
			4.00     & 1536  &      28.53    & \textbf{35.27 }& 39.36 & 47.18  &  40.60\\
			\bottomrule
		\end{tabular}%
	}
%	\vspace{-15pt}
\end{table}%

\begin{table}[t]
	\centering
	\caption{Performance under different random initialization strategies, where the constant standard deviation method performs best.}
	\label{tab:init_pref}
	\resizebox{0.99\columnwidth}{!}{
		\begin{tabular}{l|ccc|c}
			\toprule
			Initialization Method  & ARC-E & HellaSwag & C3 & Avg.   \\
			\midrule
			Constant & 37.57  & 41.16  & \textbf{49.04}  & \textbf{42.59}  \\ 
			GPT2~\cite{radford2019language} & \textbf{38.62} & 39.34  & 48.44  & 42.13  \\ 
			InternLM~\cite{team2023internlm} & 34.39 & \textbf{41.48}  & 47.70 & 41.19 \\ 
			\bottomrule
		\end{tabular}
	}
%	\vspace{-15pt}
\end{table}

In this part, we focus on the neural architecture design of LLM for edge devices by exploring the impact of depth, width and the expanding rate of Feed-Forward Networks (FFN) on the performance of a 1B-size language model. Besides accuracy on downstream tasks, decoding speed is another important aspect for tiny language models. We test the end-to-end inference speed (tokens per second) when generating 510 new tokens under two prefix tokens using a randomly initialized model. The speed is tested on a single NVIDIA V100 GPU with batch size 20 using FP16. We fix the vocabulary size to 48k as suggested in Section~\ref{Compact Tokenizer}. By constraining the model's size to 1B parameters, we explore the effects of varying the model's depth, width, and expansion rate individually. Firstly, we investigate the impact of adjusting two among the three components, while maintaining the third variable at a constant level, on the model's overall performance.

The impact of the depth and width. We thoroughly investigate representative configurations as outlined in Table~\ref{tab:with-depth-ratio}, where we can conclude that \emph{deeper tiny language models exhibit better performance, however, at the cost of inference speed.} As the depth of the model increases, the performance increases for almost all the three benchmarks. Meanwhile, we observed that when the depth is already 20, the performance improvement (41.19 $\rightarrow$ 42.02) by designing deeper architectures is minimal compared to the decrease of the inference speed (29.49 $\rightarrow$ 12.81). Therefore, we recommend setting the number of layers to around 20 for 1B-parameter model with a 48k tokenizer.

\begin{figure*}
	\centering
	\subfigure[Depth]{
		\includegraphics[width=0.27\linewidth]{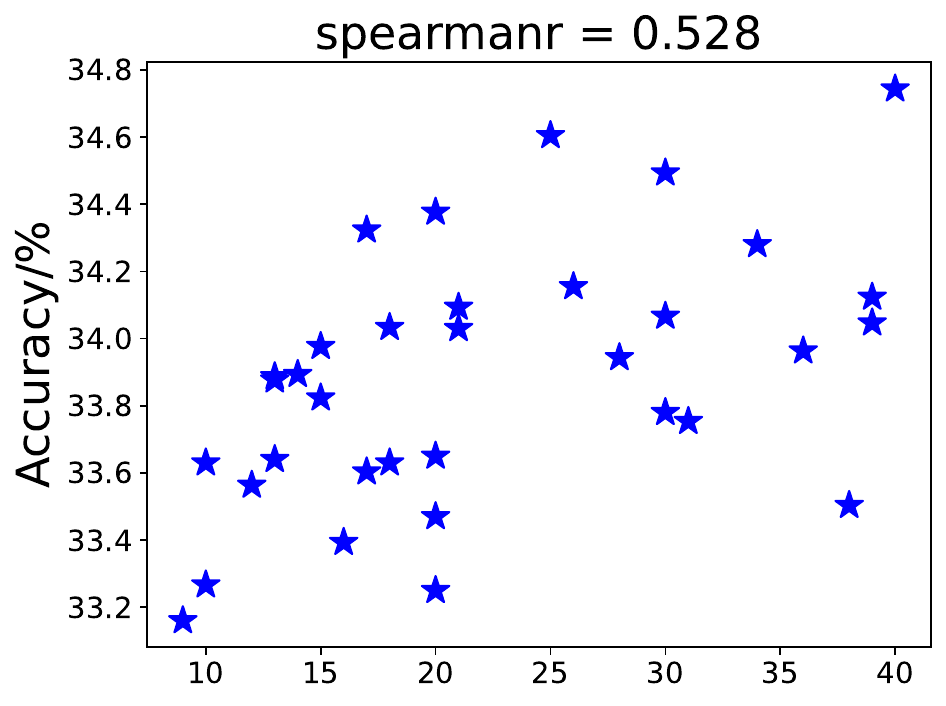}
	}
	\subfigure[Width]{
		\includegraphics[width=0.27\linewidth]{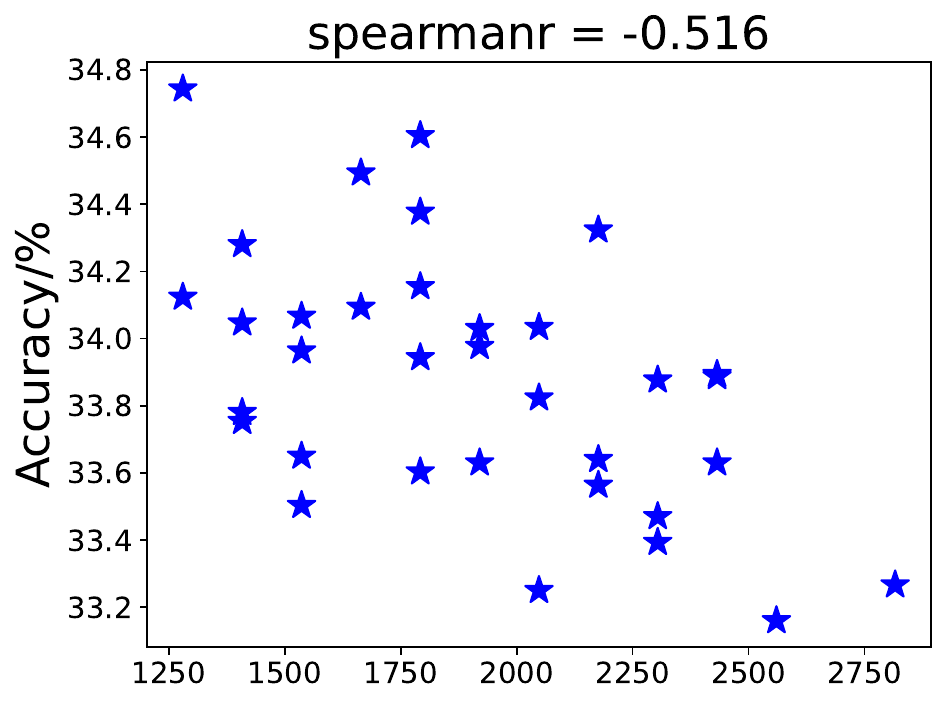}
	}
	\subfigure[Expansion rate]{
		\includegraphics[width=0.27\linewidth]{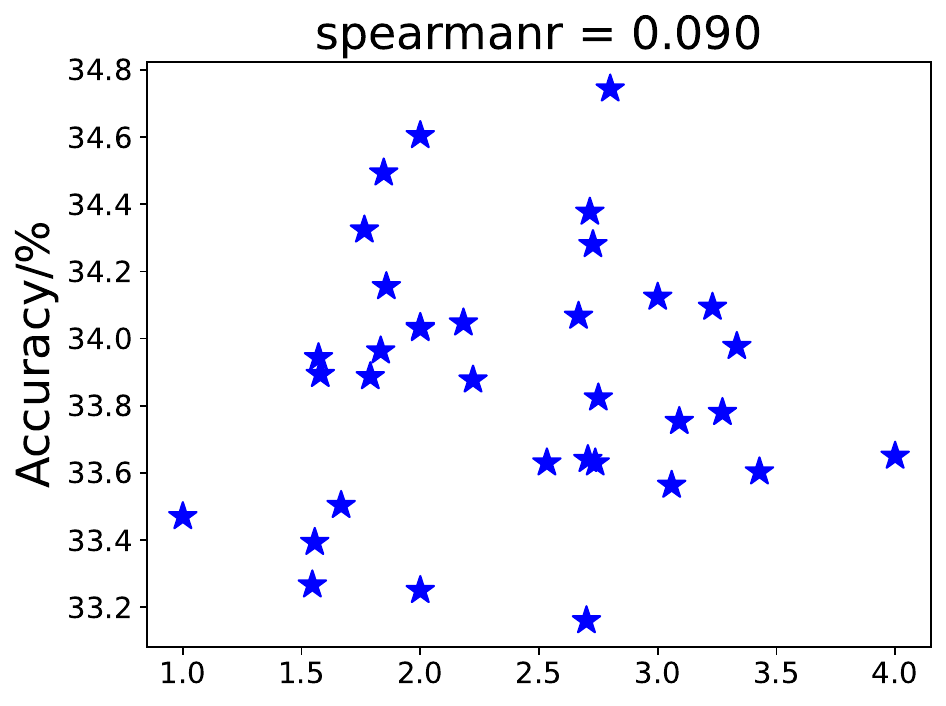}
	}
%	\vspace{-12pt}
	\caption{Performance varies \wrt~model's width, depth and expansion rate. The experiments are conducted on a streamlined dataset comprising 5B tokens. The accuracy is averaged among ARC Easy, HellaSwag and C3. Spearman coefficient is used to measure the correlation between performance and model's configure.}
	\label{img:tweak-depth-width-expansion}
%	\vspace{-15pt}
\end{figure*}

As shown in Table~\ref{tab:expanding rate}, we observe close inference speed for different expanding rates when the depth is fixed. It's obviously that the 1:1 setting gets significantly worse performance. To further investigate the interplay among depth, width and expansion rate, we sample about 30 different parameter configurations while maintaining the model size at 1B parameters and conduct training on a further streamlined dataset comprising 5B tokens. As illustrated in Figure~\ref{img:tweak-depth-width-expansion}, the correlation between the depth (width) and the downstream task's average performance is notably higher, with a Spearmanr correlation coefficient reaching up to 0.528. In contrast, there is no apparent linear relationship between the expansion rate and the model's ultimate performance.

\paragraph{Discussion.}   A compact tokenizer holds particular significance for a tiny language model, as it strikes a crucial balance between representation ability and implementation cost. The removal of low-frequency vocabularies enables the efficient elimination of substantial redundancy without significantly compromising representation capacity. Additionally, the architecture's configurations, such as width, depth, and expanding rate, exert a considerable influence on the final performance of a tiny model. Among them, depth is the primary factor for tiny language models, and deeper models usually achieve high performance at the expense of lower speed. Following the above observations, we design the architecture of PanGu-$\pi$ Pro as detailed in Table~\ref{tab:model config}.

\section{Parameter Initialization}

In this section, we investigate how to initialize model's parameters with a given neural architecture, including random initialization and inheriting parameters from a large model.  

\subsection{Random Initialization} 

When training model from scratch, the parameters are usually initialized with random numbers obeying normal distribution $N(0,\sigma^2)$ with zero mean and standard deviation $\sigma$.
A series of well-known large language models carefully design the value of $\sigma$, especially changing it \wrt layers. For example, GPT2~\cite{radford2019language} applies a scale of $1/\sqrt{N}$ to all linear layer parameters,
where $N$ is the number of residual layers.
InternLM~\cite{team2023internlm} only applies the same scale to some special linear layer parameters, namely the out projection in MHA layers and the gate projection in MLP layers.
We investigate these different initialization strategies for training tiny language model, whose results are shown in Table~\ref{tab:init_pref}.  We note that different strategies result in similar results. For simplicity and generalization, we recommend using a constant value for all layers when training tiny language models.  More analyses are presented in Appendix~\ref{sec:a-random}.

\begin{figure*}[t] 
	%\vskip 0.2in
	\begin{center}
		\centerline{\includegraphics[width=0.99\textwidth]{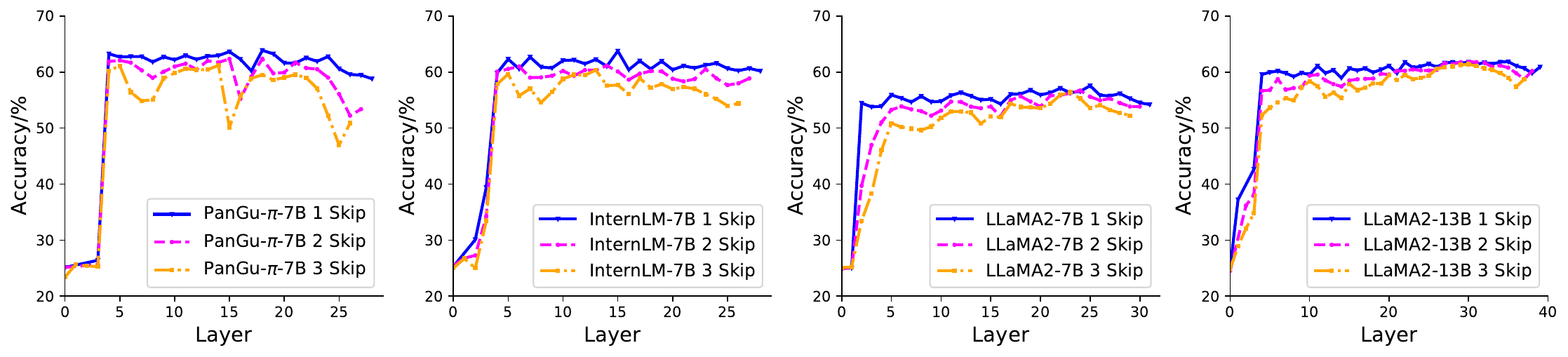}} 
%		\vspace{-15pt}
		\caption{Performance of large language models when skipping a few layers. ``$x$ Skip" denotes adjacent $x$ layers are discarded. Redundancies are observed within intermediate layers while the layers situated near the beginning and end are crucial for maintaining performance. } 
		\label{img:layer_skip}
	\end{center}
%	\vspace{-10pt}
\end{figure*}

\begin{figure}[ht]
	%\vskip 0.2in
	\begin{center}
		\centerline{\includegraphics[width=0.85\columnwidth]{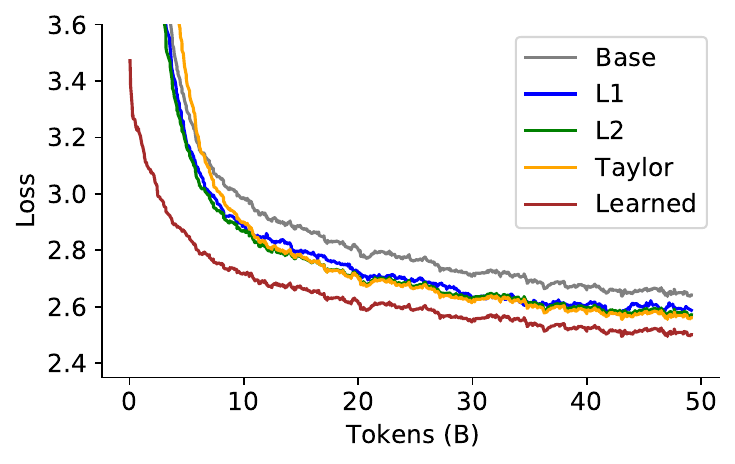}}
%		\vspace{-10pt}
		\caption{Training loss with different pruning strategies. ``Base" denotes training from scratch without inheritance. Inheriting the model parameters with pruning yields a lower loss.}
		\label{fig:prune strategy}
	\end{center}
%	\vspace{-10pt}
\end{figure}

\subsection{Parameter Inheritance} 

Besides random initialization, the initial parameter of this tiny language model can also inherit from a large language model. The strong generalization ability of large model is expected to transfer to the tiny model. Compared the tiny model, the large model  usually has more layers with more neurons. We firstly select important layers and then recognize critical neurons in the selected layers.  

\paragraph{Important layers selection.} Considering that the tiny model usually has fewer layers than the large language model, the most important layers that contribute to the final performance are required to recognize. Consequently, we conduct ablation experiments to assess the impact of individual layers on the overall performance.

To uncover general principles regarding layer importance, we conduct a variety of experiments on multiple widely-used large language models, including LLaMA2-7B, LLaMA2-13B, InternLM-7B and PanGu-$\pi$-7B. During the inference phase, we skip specific layers and assess the resulting performance drop. Three types of layer skipping experiments are conducted for each model, involving skipping one layer, two neighboring layers, and three contiguous layers. The outcomes are depicted in Figure~\ref{img:layer_skip}, where we analyze the average performance of large language models on three downstream tasks, \ie, ARC-E, HellaSwag, and C3. The $x$-axis represents the skipped layer index, while the $y$-axis signifies performance accuracy.

Some interesting common phenomenons are identified in these models. The shallow layers, especially the initial two to three layers, play a pivotal role in extracting features from input sequences. Removing these layers will incur significant performance drop on downstream tasks. Similarly, deep layers is also crucial, and removing them results in a deterioration of performance. Conversely, when the intermediate layers are removed, the performance is less affected, indicating that redundancy occurs within these layers. These layers are tend to be removed when inheriting parameters.

\paragraph{Intra-layer parameters selection.} Within a layer,  important parameters can be recognized by various metrics. How to recognizing essential  parameters has well been discovered in the model pruning area~\cite{frantar2023sparsegpt, ma2023llm}. The importance of neurons can be measured by various criteria and the most significant neurons are used as the initialization of tiny models. Weight norms, such as $\ell_1$ and $\ell_2$-norm, are commonly employed to measure importance, indicating that larger weights encapsulate more crucial information~\cite{prune_15,prune_16,prune_lamp,tang2024survey}. The first-order Taylor expansion~\cite{prune_snip,prune_synflow}, which incorporates both weight values and gradients, is regarded as a more accurate estimation of the output.  In addition to empirical criteria, essential weights can also be identified through binary masks, which are automatically learned during the training process~\cite{xia2023sheared,tang2020scop}. In the subsequent sections, we adopt these methodologies to select vital parameters from the PanGu-$\pi$-7B~\cite{wang2023PanGu} model as initial values for a 1B model with smaller weight dimensions.

\begin{table}[t]
	\centering %The data-driven learnable criteria  achieves the highest performance. 
	\caption{Comparison between different  parameter inheritance strategies. ``Base" denotes training without inheritance.}
	\label{tab:prune strategy}%
	%\vskip 0.15in
	\resizebox{0.99\columnwidth}{!}{
		\begin{tabular}{l|ccc|c}
			\toprule
			Inheritance Strategy & ARC-E & HellaSwag   & C3  & Avg.  \\
			\midrule
			Base & 36.68 & 40.34 & 49.15  & 42.06 \\
			L1~\cite{ma2023llm}     & 39.51 & 47.70 & 50.96  & 46.06 \\
			L2~\cite{ma2023llm}     & 41.98 & 48.33   & 50.68  & 47.00 \\
			Taylor~\cite{ma2023llm} & \textbf{43.21} & 48.43    & \textbf{52.05} & 47.90 \\
			Learned~\cite{xia2023sheared,tang2020scop} & 40.74 & \textbf{51.77}  & 51.73 & \textbf{48.08} \\
			\bottomrule
		\end{tabular}%
	}
	%\vspace{-20pt}
\end{table}%

Training loss curves and evaluation results are presented in Figure~\ref{fig:prune strategy} and Table~\ref{tab:prune strategy}. In comparison to the baseline model initialized randomly, each of the small models initialized with the pruning strategy converges to a lower loss. Among the empirical criteria, the Taylor expansion yields superior results, primarily attributed to its accurate estimation of neuron importance. The model pruned using learnable masks starts with a significantly lower initial loss than the other models and ultimately converging to the lowest loss. Evaluation results across the three datasets validate the effectiveness of parameter inheritance. We recommend inheriting model parameters with the learnable strategies.

\paragraph{Discussion.} The aforementioned observation confirms that the initial parameters of a model exert a substantial influence on both the convergence rate and ultimate performance of  a tiny language model. Opting to inherit parameters from a larger model is generally a more favorable choice, as it allows the smaller model to assimilate the robust representation abilities of its larger models. The process of selecting significant parameters is a crucial step in this regard. Through thorough empirical investigation, we have observed that intermediate layers tend to exhibit more redundancy, and  data-driven learnable masks proves effective in excavating these redundant parameters.

\begin{figure}
	\centering
	\subfigure[$r=0.5$]{
		\includegraphics[width=0.46\linewidth]{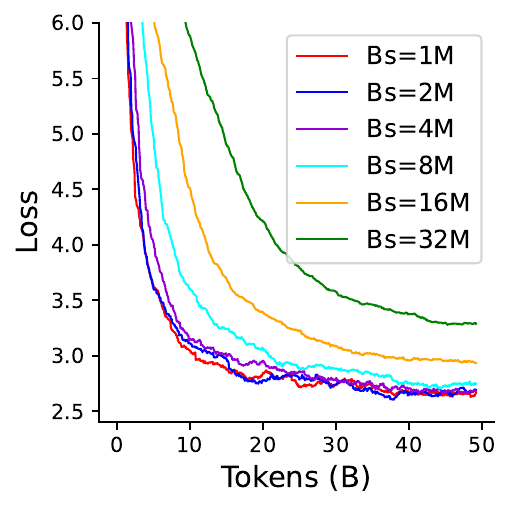}
	}
	\subfigure[$r=1.0$]{
		\includegraphics[width=0.46\linewidth]{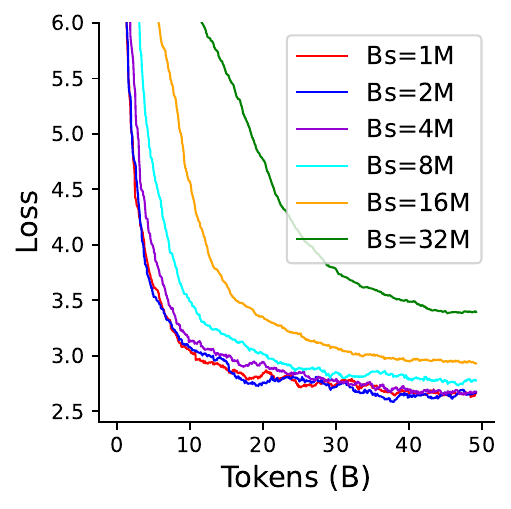}
	}
%	\vspace{-10pt}
	\caption{Training losses under different batchsize \& learning rate.}
	\label{img: bs_lr_loss}
%	\vspace{-15pt}
\end{figure}

\section{Model Optimization}
In this section, we investigate the optimization strategies with given neural architecture and initial parameters. Firstly, the scaling rule between learning rate and batch size is analyzed. Besides, we observe substantial performance improvement from continuing multiple rounds of training.

\subsection{Batchsize \& Learning Rate}

In practical language model training, the choice of batch size is frequently tailored to the computational resources at hand. When dealing with a limited number of GPUs, opting for a smaller batch size becomes necessary. Conversely, in scenarios where a substantial number of GPUs is at our disposal, enlarging the batch size can effectively diminish the number of iterations, thereby expediting the overall training process. 

\begin{figure}[t]
	\begin{center}
		\vspace{6pt}
		\centerline{\includegraphics[width=0.8\columnwidth]{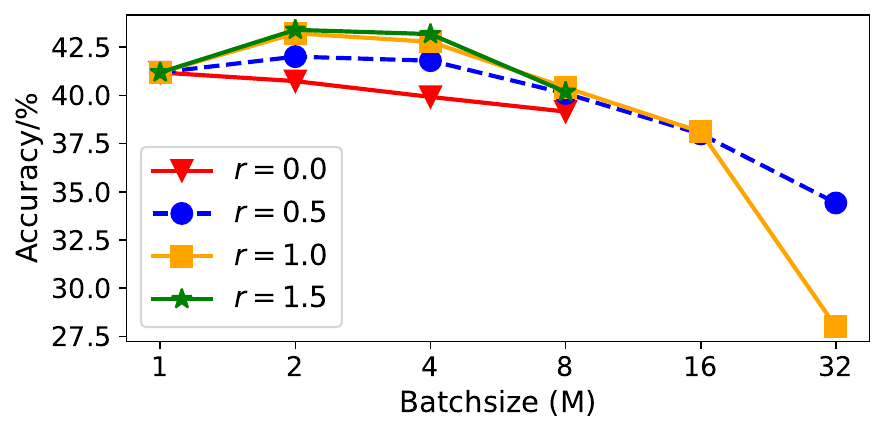}}
%		\vspace{-10pt}
		\caption{Performance under different batchsize \& learning rate.} 
		\label{img: bs_lr_perf}
	\end{center}
%	\vspace{-15pt}
\end{figure}

However, adjusting the batch size typically has a notable impact on the final performance of the model. When increasing the batch size, it is common for the learning rate to be adjusted proportionally. We explore their combined effects in Figure~\ref{img: bs_lr_loss} and Figure~\ref{img: bs_lr_perf}, using the formula $lr = (bs/bs_0)^r \times lr_0$, where the default batchsize $bs_0$ and learning rate $lr_0$ are set to 1M and $1\times10^{-4}$, respectively. $r$ denotes the increment rate, which is usually set as 0.5 or 1.0~\cite{krizhevsky2014one, goyal2017accurate}. When the batchsize is smaller than 4M, the convergence speeds with different learning rates  remain consistent. When the batchsize further increases, a moderate increment rate ($r=0.5$) is preferable. With the same training tokens, employing an excessively large batchsize ($\geq$16M) adversely affects the convergence speed. In the majority of cases, a batch size smaller than 4M is considered the safe range for optimizing model performance. Otherwise,  optimization strategies need to be specifically tailored for large batch sizes~\cite{keskar2016large,you2017large,you2019large}.

\subsection{Multiple-Round Training} 

The existing methods usually train the language model with only one round, \ie, all the data are only used for one time to update the model, leaving the model's parameters unconverged. Besides, learning on large corpora may suffer from the catastrophic forgetting~\cite{toneva2018empirical, winata2023overcoming} issue, \ie, the model performance drops for data seen before. For tiny models, the limited model capacity makes the forgetting problem more serious. Continuing training the model can further reduce the training loss. 

\begin{figure}[t] 
	\begin{center}
		\centerline{\includegraphics[width=0.8\columnwidth]{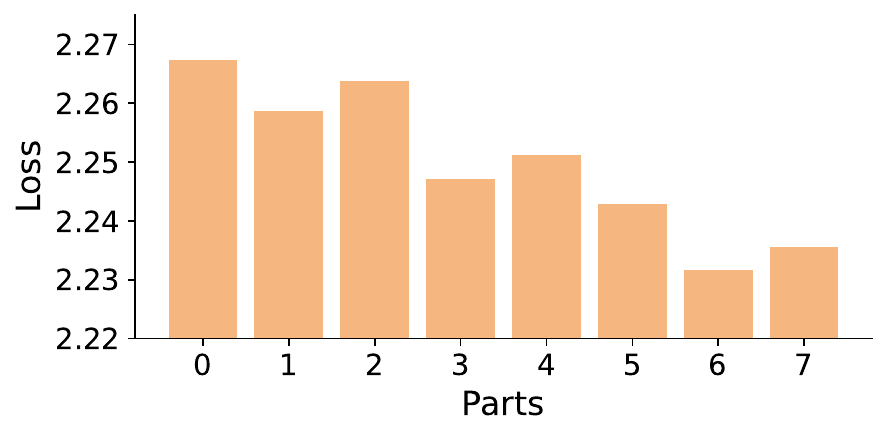}}
%		\vspace{-10pt}
		\caption{Loss value varies \wrt~data on different iterations using a pretrained PanGu-$\pi$-1B model. The loss is averaged among batches in each part.}
		\label{img:prevloss}
	\end{center}
%	\vspace{-20pt}
\end{figure}

We conduct a simple experiment to validate the forgetting problem. As the training loss is calculated by the model parameters at the corresponding timestamp and the model parameters are updated as the training continues, the later data tend to have low loss values. Therefore, We recompute the batch-wise loss on the previous data using a PanGu-$\pi$-1B model trained on 1.6T tokens. The training data is evenly and randomly divided into eight parts before training. Figure~\ref{img:prevloss} shows how loss value varies \wrt~data on each part. The high loss indicate previous knowledge have been seriously forgot. Therefore, it is necessary to train the model for multiple rounds to fit the forgotten data.

\begin{figure}[t]
	%\vskip 0.2in
	\begin{center}
		\centerline{\includegraphics[width=0.8\columnwidth]{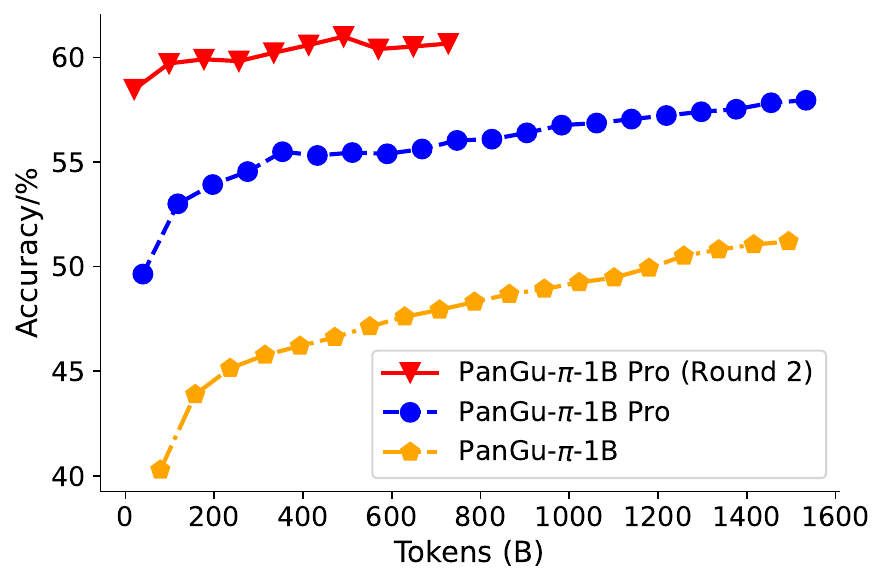}}
%		\vspace{-10pt}
		\caption{Accuracies of PanGu-$\pi$-1B and PanGu-$\pi$-1B Pro on HellaSwag during training.  }
		\label{evolves}
	\end{center}
%	\vspace{-15pt}
\end{figure}

To reduce the training cost, we propose a simple data refining strategy for the multiple-round training. Considering some examples are hard to fit, they should be used for further training with a high probability. Denoting the loss values in certain part as $L=\{l_1,\l_2,\cdots,l_N\}$, where $N$ is the total batches in this part. Note that data are randomly shuffled in the training process, and thus each batch contains various type data. In each part, the
loss values are normalized, denoting the sampling probability, \ie, $p_i = \frac{\exp(l_i)}{\sum_{j=1}^N \exp(l_j)}$. In the next round training we sample $N_0$ batches out of $N$ according to the sampling probability $p$. The impact of sampling rate ($r = \frac{N_0}{N}$) is shown in Table~\ref{tab: sr-multi}. It shows that a higher sampling rate tends to achieve high performance. The performance improvement is marginal when the sampling rate $r$ exceeds 50\%. We plot how the evaluation metric on HellaSwag evolves during training in Figure~\ref{evolves}. As the second-round training goes ahead, the accuracy on HellaSwag keeps rising but get converged in the later phase. In Table~\ref{tab: Multi-round}, we also try to train the models with more rounds. However, the performance also saturate gradually. To achieve a balance between performance and training efficiency, we recommend to train the model with two rounds and  set  sampling rate to 50\%.

\paragraph{Discussion.} 
In contrast to larger models, tiny language models face a significant challenge of data forgetting due to their limited capacity. As a result, adopting a multi-round training approach becomes crucial to enhance performance. Employing data sampling proves effective in improving learning on challenging examples while simultaneously reducing training costs. Additionally, the choice of batch size and learning rate plays a pivotal role in model performance. For optimal results, it is advisable to use a batch size smaller than 4M, unless a specialized large batch optimizer is employed for a tailored approach.

\begin{table}[t]
	\centering
	\caption{Sampling rate for the next round training. The model is training with two rounds. $r=0$ denotes training with one round.}
	\label{tab: sr-multi}%
	\resizebox{0.99\columnwidth}{!}{
		\begin{tabular}{c|ccc|c}
			\toprule
			Sampling Rate $r$   & ARC-E & HellaSwag  & C3& Average \\
			\midrule
			0\%           & 42.68 & 57.95 & 54.19 & 51.61  \\
			25\%        & 43.95 &  59.55  & 56.01 &  53.17  \\
			50\%         & 45.33 & 60.67  & 57.37 & 54.46  \\
			75\%         & 45.52 &  60.34  & 58.16 & 54.67  \\
			100\%       & 44.98 & 60.88  & 58.74 &  54.87 \\
			\bottomrule
		\end{tabular}%
	}
%	\vspace{-15pt}
\end{table}%

\begin{table}[t]
	\centering
	\caption{The impact of number of training rounds. The sampling rate $r$ is set to 50\%.}
	\label{tab: Multi-round}%
	\vskip 0.05in
	\resizebox{0.99\columnwidth}{!}{
		\begin{tabular}{c|ccc|c}
			\toprule
			Training round  & ARC-E & HellaSwag  & C3& Average \\
			\midrule
			Single round      & 42.68 & 57.95 & 54.19 & 51.61   \\
			Two round         & 45.33 & 60.67  & 57.37 & 54.46   \\
			Three round       & 45.11 & 61.32  & 56.88 &  54.44\\
			\bottomrule
		\end{tabular}%
	}
%	\vspace{-15pt}
\end{table}%

\begin{table*}[htbp]
	\centering
	\caption{Comparison with SOTA open-source tiny language models. The best model is listed in bold and second-best is listed in underlined.}
	\label{tab:benchmark}%
	\resizebox{0.98\textwidth}{!}{
		\begin{tabular}{lccccccccccc}
			\toprule
			& \multicolumn{4}{c}{\textbf{Examination}} & \textbf{Knowledge} & \multicolumn{2}{c}{\textbf{Reasoning}} & \multicolumn{3}{c}{\textbf{Understanding}} & \multirow{2}[4]{*}{\textbf{Average}} \\
			\cmidrule(l{0.5em}r{0.5em}){2-5} \cmidrule(l{0.5em}r{0.5em}){6-6} \cmidrule(l{0.5em}r{0.5em}){7-8} \cmidrule(l{0.5em}r{0.5em}){9-11}  \textbf{Models} & C-Eval & CMMLU & MMLU  & AGI-Eval & BoolQ & AX-b  & PIQA  & EPRSTMT & XSum  & C3    &  \\
			\midrule
			MobileLLaMA-1.4B & 23.93  & 25.10  & 25.05  & 18.53  & 58.75  & 45.20  & 71.27  & 46.25  & 18.19  & 37.42  & 36.97  \\
			Sheared-LLaMA-1.3B & 24.28  & 25.10  & 25.77  & 18.01  & 62.39  & 43.57  & 72.91  & 46.25  & 16.44  & 35.45  & 37.02  \\
			TinyLLaMA-1.1B & 27.85  & 24.64  & 25.75  & 18.54  & 56.06  & 45.47  & 70.62  & 46.25  & 20.15  & 36.71  & 37.20  \\
			MobileLLaMA-2.7B & 23.53  & 25.55  & 26.63  & 18.43  & 54.74  & {55.80}  & 72.85  & 46.25  & 16.96  & 36.11  & 37.69  \\
			Chinese-LLaMA2-1.3B & 28.70  & 24.78  & 24.55  & 19.40  & 56.79  & 47.46  & 56.91  & 72.50  & 8.90  & 43.12  & 38.31  \\
			RWKV-5-1.5B & 25.92  & 25.14  & 25.66  & 19.01  & 62.29  & 54.05  & 71.22  & 46.25  & {20.67}  & 49.15  & 39.94  \\
			Phi-1.3B & 27.78  & 25.85  & 44.32  & 23.42  & {73.52}  & 44.20  & 76.99  & 50.00  & 14.90  & 38.96  & 41.99  \\
			PanGu-$\pi$-1B & 36.85  & 35.90  & 35.96  & 30.77  & 58.44  & 43.48  & 61.92  & 55.62  & 15.92  & 49.21  & 42.41  \\
			Open-LLaMA-3B & 27.50  & 25.42  & 27.09  & 20.68  & 60.58  & 52.72  & {77.09}  & 82.50  & 19.75  & 43.23  & 43.66  \\
			Phi2-2.7B & 31.86  & 32.18  & {58.49} & 28.51  & \underline{77.40} & 43.57  & {78.89} & 46.25  & 13.66  & 40.11  & 45.09  \\
			Gemma-2.5B & 31.34 & 31.08 & 41.92 & 24.27 & 62.13 & 51.89 & 78.16 & 55.61 & 19.25 & 50.38 & 44.60  \\ 
			PanGu-$\pi$-1B Pro ({Ours}) & 46.50  & 46.56  & 50.38  & {41.58} & 63.43  & 53.99  & 64.96  & 74.38  & 18.40  & 52.66  & 51.28  \\
			Qwen-1.8B & {53.60} & {52.12} & 46.43  &{35.83}  & 64.31  & {57.79} & 73.83  & \underline{88.12}  & 20.03  & {58.30}  & {55.04}  \\			
			Qwen2-1.5B & \textbf{70.66} & \textbf{70.63} & 57.49 & 40.31 & 73.15 & 50.22 & 69.57 & 82.14 & 13.78 & \textbf{76.80} & 60.47  \\
			PanGu-$\pi$-1.5B Pro ({Ours}) & {52.39}  & {48.51}  & {62.54}  & \underline{44.89}  & 70.61 & \textbf{67.93}  & \underline{79.55}  & {86.12} & \textbf{24.61} & {69.24} & {60.64}  \\
			Qwen2.5-1.5B & \underline{68.01} & {64.96} &  \underline{62.23} & {42.75} & 76.98 & 55.07 & 74.24 & 85.09 & 13.02 &  \underline{75.81} &\underline{61.81} \\
			PanGu-$\pi$-1.5B Pro* ({Ours})\tablefootnote{The performance is further improved by training model with 6T data.} & 65.93 & \underline{68.82} & \textbf{64.26} & \textbf{47.94} & \textbf{80.85} & \underline{63.37} & \textbf{80.08} &\textbf{90.21} & \underline{23.49} & 69.36 & \textbf{65.43}  \\
			\bottomrule
		\end{tabular}%
	}
	%    \vspace{-3pt}
\end{table*}%

\section{PanGu-$\pi$ Pro}
\label{V2}

Based on the above extensive and rigorous set of experiments, we make a significant improvement on our previous PanGu-$\pi$-1B and meanwhile construct a larger and more powerful PanGu-$\pi$-1.5B Pro. In this section, we make a comparison with the existing open-source tiny language models, where our result establishes a new SOTA. Specifically, PanGu-$\pi$-1.5B Pro outperforms the recent proposed Qwen-1.8B~\cite{qwen} and Phi2-2.7B~\cite{textbooks2}, which is 1.8x larger, in average performance.

\paragraph{Implementation details.} The pre-training data, which consists of 1.6T tokens, is gathered from diverse sources from the Internet, covering English and Chinese corpus with around $1:1$ scale. The used 48k tokenizer is built by byte-pair encoding (BPE, ~\citet{shibata1999byte}) from SentencePiece~\cite{kudo2018sentencepiece} upon our data. Our models are trained using the AdamW optimizer~\cite{loshchilov2017decoupled} with $\beta_1 = 0.9, \beta_2 = 0.95$ utilizing the cosine learning rate decay~\cite{loshchilov2016sgdr} with an initial learning rate $2 \times 10^{-4}$. The total batch size for the training process is 2M. We follow the PanGu-$\pi$~\cite{wang2023PanGu} architecture while making PanGu-$\pi$-1B much deeper. The detailed configuration can be found in Table~\ref{tab:model config}. We set the expansion rate to 2.77 as suggested in Table~\ref{tab:expanding rate}. For the parameters initialization method, we inherit parameters from PanGu-$\pi$-7B via learnable binary mask after removing 
intermediate redundant layers. We use the Huawei Ascend 910 card to train and evaluate the proposed PanGu-$\pi$ Pro.

\begin{table}[h]
	\centering
%	\vspace{-10pt}
	\caption{Model configuration.}
	\label{tab:model config}%
	\resizebox{0.98\columnwidth}{!}{
		\begin{tabular}{ccccc}
			\toprule
			Model & Width &  Depth  & Vocabulary & Initialization \\
			\midrule
			PanGu-$\pi$-1B & 2048   & 12 & 100883 & Random \\
			PanGu-$\pi$-1B Pro  & 1792  & 21  & 48000 &  PanGu-$\pi$-7B \\
			PanGu-$\pi$-1.5B Pro & 2048   & 22   & 48000 &  PanGu-$\pi$-7B\\
			\bottomrule
		\end{tabular}%
	}
%	\vspace{-2pt}
\end{table}%

\paragraph{Benchmarks.} We use OpenCompass~\cite{2023opencompass} to evaluate on an extensive suite of downstream tasks, covering {examination}, knowledg, {reasoning}, and {understanding} abilities for a comprehensive comparison. C-Eval~\citep{huang2023ceval} is a Chinese benchmark to evaluate the knowledge and reasoning abilities. CMMLU~\cite{li2023cmmlu} covers 67 topics including science, engineering, and humanities. MMLU~\cite{hendryckstest2021} proposes an English benchmark for measuring LLM's multitask accuracy by covering 57 tasks including mathematics, history, computer science, and law. AGI-Eval~\cite{zhong2023agieval} is a benchmark specifically designed to evaluate the general abilities in tasks pertinent to human cognition and problem-solving. BoolQ~\cite{clark2019boolq} is a reading comprehension dataset to evaluate the difficult entailment-like inference ability of LLMs. AX-b~\cite{wang2020superglue} is a broad-coverage diagnostic task and PIQA~\cite{Bisk2020} is a physical interaction question-answering task. EPRSTM~\cite{xu2021fewclue} is a binary sentiment analysis dataset based on product reviews. XSum~\cite{narayan2018dont} is a summarization task collected from the British Broadcasting Corporation and C3~\cite{sun2019investigating} contains 13,369 documents and their associated 19,577 multiple-choice questions.

\paragraph{Comparison with tiny language models.} We collect multiple tiny language models with different sizes, ranging from 1B to 3B. These include TinyLLaMA-1.1B~\cite{tinyllama}, Chinese-LLaMA2-1.3B~\cite{Chinese-LLaMA-Alpaca}, Sheared-LLaMA-1.3B~\cite{xia2023sheared}, and Open-LLaMA-3B~\cite{openlm2023openllama}. Meituan~\cite{chu2023mobilevlm} released
MobileLLaMA-1.4B and MobileLLaMA-2.7B that were trained from scratch on the RedPajama dataset~\cite{together2023redpajama}. Microsoft developed the series of Phi~\cite{textbooks2} that focusing on using ``textbook-quality" data with small language models. RWKV-5-1.5B~\cite{peng2023rwkv} is a parallelizable RNN with Transformer-level LLM Performance. Qwen-1.8B~\cite{qwen} is pretrained on 2.2 trillion tokens including web texts, books, codes, \etc.

Extensive experiments in Table~\ref{tab:benchmark} show that PanGu-$\pi$-1.5B Pro significantly outperforms existing LLMs of similar or even larger sizes, \eg, Phi2-2.7B and Open-LLaMA-3B. We observe a notable improvement of 8.77 on average performance from PanGu-$\pi$-1B to PanGu-$\pi$-1B Pro. With 16.67\% fewer parameters, PanGu-$\pi$-1.5B Pro outperforms Qwen-1.8B~\cite{qwen}  and exhibits the best or second-best performance in the vast majority of the benchmarks. Overall, our model exhibits consistently better average performance compared to the current state-of-the-art models.

From PanGu-$\pi$-1B, we gradually add the core components to validate the effectiveness of our methodology. As shown in Figure~\ref{img:overall}, the removal of low-frequency vocabularies leads to an improvement of average performance from 42.41 to 44.11 while the architecture tweak contributes another 2.42 improvement. Parameter inheritance, the most effective approach, pushes the average performance to 49.79. Multiple-round training further enhances the average performance of PanGu-$\pi$-1B Pro.

\section{Conclusion and Discussion}

In this paper, we systematically discuss how to construct a tiny language model from three perspectives, \ie, neural architecture, parameter initialization, and optimization strategies. By carefully designed empirical study, we recognized several effective design formulas to improve performance  with given  parameter restriction and data size, including compact tokenizer, architecture tweak, parameter inheritance, multiple-round training \etc. Then we train PanGu-$\pi$ Pro models with 1B and 1.5B parameters, which significantly improve performance than the baseline models.

Based on the observations, we also note several intriguing directions for further exploration.  In terms of neural architecture, how to directly learn a compact tokenizer that seamlessly integrates both representation ability and parameter efficiency. Additionally, exploring hardware-friendly architectures holds promise for mitigating computational and storage costs.  For example, GQA~\cite{ainslie2023gqa} is an effective strategy to reduce RAM require of edge devices~(refer to Appendix~\ref{convert}). Concerning model optimization, the importance of effective parameter initialization cannot be overstated, setting the model on a path toward high performance from the outset. Nevertheless, the challenge remains in identifying effective parameters, which is an open question in the field. Besides, the training characteristics of tiny models differ significantly from their larger counterparts.  For instance, within the framework of multiple-round training, there is an urgent demand for the development of new parameter optimization techniques and data refining methods.  Numerous questions warrant in-depth exploration, and we hope the findings presented in this paper can spark inspiration for further research.

\section*{Impact Statement} 
This paper presents work whose goal is to advance the field of Machine Learning. There are many potential societal consequences of our work, none which we feel must be specifically highlighted here.

\bibliography{ref}

\begin{thebibliography}{58}
\providecommand{\natexlab}[1]{#1}
\providecommand{\url}[1]{\texttt{#1}}
\expandafter\ifx\csname urlstyle\endcsname\relax
  \providecommand{\doi}[1]{doi: #1}\else
  \providecommand{\doi}{doi: \begingroup \urlstyle{rm}\Url}\fi

\bibitem[Achiam et~al.(2023)Achiam, Adler, Agarwal, Ahmad, Akkaya, Aleman,
  Almeida, Altenschmidt, Altman, Anadkat, et~al.]{achiam2023gpt}
Achiam, J., Adler, S., Agarwal, S., Ahmad, L., Akkaya, I., Aleman, F.~L.,
  Almeida, D., Altenschmidt, J., Altman, S., Anadkat, S., et~al.
\newblock Gpt-4 technical report.
\newblock \emph{arXiv preprint arXiv:2303.08774}, 2023.

\bibitem[Ainslie et~al.(2023)Ainslie, Lee-Thorp, de~Jong, Zemlyanskiy,
  Lebr{\'o}n, and Sanghai]{ainslie2023gqa}
Ainslie, J., Lee-Thorp, J., de~Jong, M., Zemlyanskiy, Y., Lebr{\'o}n, F., and
  Sanghai, S.
\newblock Gqa: Training generalized multi-query transformer models from
  multi-head checkpoints.
\newblock \emph{arXiv preprint arXiv:2305.13245}, 2023.

\bibitem[Bai et~al.(2023)Bai, Bai, Chu, Cui, Dang, Deng, Fan, Ge, Han, Huang,
  Hui, Ji, Li, Lin, Lin, Liu, Liu, Lu, Lu, Ma, Men, Ren, Ren, Tan, Tan, Tu,
  Wang, Wang, Wang, Wu, Xu, Xu, Yang, Yang, Yang, Yang, Yao, Yu, Yuan, Yuan,
  Zhang, Zhang, Zhang, Zhang, Zhou, Zhou, Zhou, and Zhu]{qwen}
Bai, J., Bai, S., Chu, Y., Cui, Z., Dang, K., Deng, X., Fan, Y., Ge, W., Han,
  Y., Huang, F., Hui, B., Ji, L., Li, M., Lin, J., Lin, R., Liu, D., Liu, G.,
  Lu, C., Lu, K., Ma, J., Men, R., Ren, X., Ren, X., Tan, C., Tan, S., Tu, J.,
  Wang, P., Wang, S., Wang, W., Wu, S., Xu, B., Xu, J., Yang, A., Yang, H.,
  Yang, J., Yang, S., Yao, Y., Yu, B., Yuan, H., Yuan, Z., Zhang, J., Zhang,
  X., Zhang, Y., Zhang, Z., Zhou, C., Zhou, J., Zhou, X., and Zhu, T.
\newblock Qwen technical report.
\newblock \emph{arXiv preprint arXiv:2309.16609}, 2023.

\bibitem[Bisk et~al.(2020)Bisk, Zellers, Bras, Gao, and Choi]{Bisk2020}
Bisk, Y., Zellers, R., Bras, R.~L., Gao, J., and Choi, Y.
\newblock Piqa: Reasoning about physical commonsense in natural language.
\newblock In \emph{Thirty-Fourth AAAI Conference on Artificial Intelligence},
  2020.

\bibitem[Brown et~al.(2020)Brown, Mann, Ryder, Subbiah, Kaplan, Dhariwal,
  Neelakantan, Shyam, Sastry, Askell, et~al.]{brown2020language}
Brown, T., Mann, B., Ryder, N., Subbiah, M., Kaplan, J.~D., Dhariwal, P.,
  Neelakantan, A., Shyam, P., Sastry, G., Askell, A., et~al.
\newblock Language models are few-shot learners.
\newblock \emph{Advances in neural information processing systems},
  33:\penalty0 1877--1901, 2020.

\bibitem[Chowdhery et~al.(2023)Chowdhery, Narang, Devlin, Bosma, Mishra,
  Roberts, Barham, Chung, Sutton, Gehrmann, et~al.]{chowdhery2023palm}
Chowdhery, A., Narang, S., Devlin, J., Bosma, M., Mishra, G., Roberts, A.,
  Barham, P., Chung, H.~W., Sutton, C., Gehrmann, S., et~al.
\newblock Palm: Scaling language modeling with pathways.
\newblock \emph{Journal of Machine Learning Research}, 24\penalty0
  (240):\penalty0 1--113, 2023.

\bibitem[Chu et~al.(2023)Chu, Qiao, Lin, Xu, Yang, Hu, Wei, Zhang, Zhang, Wei,
  and Shen]{chu2023mobilevlm}
Chu, X., Qiao, L., Lin, X., Xu, S., Yang, Y., Hu, Y., Wei, F., Zhang, X.,
  Zhang, B., Wei, X., and Shen, C.
\newblock Mobilevlm: A fast, strong and open vision language assistant for
  mobile devices, 2023.

\bibitem[Clark et~al.(2019)Clark, Lee, Chang, Kwiatkowski, Collins, and
  Toutanova]{clark2019boolq}
Clark, C., Lee, K., Chang, M.-W., Kwiatkowski, T., Collins, M., and Toutanova,
  K.
\newblock Boolq: Exploring the surprising difficulty of natural yes/no
  questions.
\newblock In \emph{NAACL}, 2019.

\bibitem[Clark et~al.(2018)Clark, Cowhey, Etzioni, Khot, Sabharwal, Schoenick,
  and Tafjord]{clark2018think}
Clark, P., Cowhey, I., Etzioni, O., Khot, T., Sabharwal, A., Schoenick, C., and
  Tafjord, O.
\newblock Think you have solved question answering? try arc, the ai2 reasoning
  challenge, 2018.

\bibitem[Computer(2023)]{together2023redpajama}
Computer, T.
\newblock Redpajama: An open source recipe to reproduce llama training dataset,
  2023.
\newblock URL \url{https://github.com/togethercomputer/RedPajama-Data}.

\bibitem[Contributors(2023)]{2023opencompass}
Contributors, O.
\newblock Opencompass: A universal evaluation platform for foundation models.
\newblock \url{https://github.com/open-compass/opencompass}, 2023.

\bibitem[Cui et~al.(2023)Cui, Yang, and Yao]{Chinese-LLaMA-Alpaca}
Cui, Y., Yang, Z., and Yao, X.
\newblock Efficient and effective text encoding for chinese llama and alpaca.
\newblock \emph{arXiv preprint arXiv:2304.08177}, 2023.
\newblock URL \url{https://arxiv.org/abs/2304.08177}.

\bibitem[Frantar \& Alistarh(2023)Frantar and Alistarh]{frantar2023sparsegpt}
Frantar, E. and Alistarh, D.
\newblock Sparsegpt: Massive language models can be accurately pruned in
  one-shot.
\newblock In \emph{International Conference on Machine Learning}, pp.\
  10323--10337. PMLR, 2023.

\bibitem[Geng \& Liu(2023)Geng and Liu]{openlm2023openllama}
Geng, X. and Liu, H.
\newblock Openllama: An open reproduction of llama, May 2023.
\newblock URL \url{https://github.com/openlm-research/open_llama}.

\bibitem[Goyal et~al.(2017)Goyal, Doll{\'a}r, Girshick, Noordhuis, Wesolowski,
  Kyrola, Tulloch, Jia, and He]{goyal2017accurate}
Goyal, P., Doll{\'a}r, P., Girshick, R., Noordhuis, P., Wesolowski, L., Kyrola,
  A., Tulloch, A., Jia, Y., and He, K.
\newblock Accurate, large minibatch sgd: Training imagenet in 1 hour.
\newblock \emph{arXiv preprint arXiv:1706.02677}, 2017.

\bibitem[Guo et~al.(2016)Guo, Yao, and Chen]{prune_16}
Guo, Y., Yao, A., and Chen, Y.
\newblock Dynamic network surgery for efficient dnns.
\newblock In Lee, D.~D., Sugiyama, M., von Luxburg, U., Guyon, I., and Garnett,
  R. (eds.), \emph{Advances in Neural Information Processing Systems 29: Annual
  Conference on Neural Information Processing Systems 2016, December 5-10,
  2016, Barcelona, Spain}, pp.\  1379--1387, 2016.

\bibitem[Han et~al.(2015)Han, Pool, Tran, and Dally]{prune_15}
Han, S., Pool, J., Tran, J., and Dally, W.~J.
\newblock Learning both weights and connections for efficient neural networks.
\newblock \emph{CoRR}, abs/1506.02626, 2015.
\newblock URL \url{http://arxiv.org/abs/1506.02626}.

\bibitem[Hendrycks et~al.(2021)Hendrycks, Burns, Basart, Zou, Mazeika, Song,
  and Steinhardt]{hendryckstest2021}
Hendrycks, D., Burns, C., Basart, S., Zou, A., Mazeika, M., Song, D., and
  Steinhardt, J.
\newblock Measuring massive multitask language understanding.
\newblock \emph{Proceedings of the International Conference on Learning
  Representations (ICLR)}, 2021.

\bibitem[Huang et~al.(2023)Huang, Bai, Zhu, Zhang, Zhang, Su, Liu, Lv, Zhang,
  Lei, Fu, Sun, and He]{huang2023ceval}
Huang, Y., Bai, Y., Zhu, Z., Zhang, J., Zhang, J., Su, T., Liu, J., Lv, C.,
  Zhang, Y., Lei, J., Fu, Y., Sun, M., and He, J.
\newblock C-eval: A multi-level multi-discipline chinese evaluation suite for
  foundation models.
\newblock \emph{arXiv preprint arXiv:2305.08322}, 2023.

\bibitem[Keskar et~al.(2016)Keskar, Mudigere, Nocedal, Smelyanskiy, and
  Tang]{keskar2016large}
Keskar, N.~S., Mudigere, D., Nocedal, J., Smelyanskiy, M., and Tang, P. T.~P.
\newblock On large-batch training for deep learning: Generalization gap and
  sharp minima.
\newblock \emph{arXiv preprint arXiv:1609.04836}, 2016.

\bibitem[Krizhevsky(2014)]{krizhevsky2014one}
Krizhevsky, A.
\newblock One weird trick for parallelizing convolutional neural networks.
\newblock \emph{arXiv preprint arXiv:1404.5997}, 2014.

\bibitem[Kudo \& Richardson(2018)Kudo and Richardson]{kudo2018sentencepiece}
Kudo, T. and Richardson, J.
\newblock Sentencepiece: A simple and language independent subword tokenizer
  and detokenizer for neural text processing.
\newblock \emph{arXiv preprint arXiv:1808.06226}, 2018.

\bibitem[Lee et~al.(2021)Lee, Park, Mo, Ahn, and Shin]{prune_lamp}
Lee, J., Park, S., Mo, S., Ahn, S., and Shin, J.
\newblock Layer-adaptive sparsity for the magnitude-based pruning.
\newblock In \emph{9th International Conference on Learning Representations,
  {ICLR} 2021, Virtual Event, Austria, May 3-7, 2021}. OpenReview.net, 2021.
\newblock URL \url{https://openreview.net/forum?id=H6ATjJ0TKdf}.

\bibitem[Lee et~al.(2019)Lee, Ajanthan, and Torr]{prune_snip}
Lee, N., Ajanthan, T., and Torr, P. H.~S.
\newblock Snip: single-shot network pruning based on connection sensitivity.
\newblock In \emph{7th International Conference on Learning Representations,
  {ICLR} 2019, New Orleans, LA, USA, May 6-9, 2019}. OpenReview.net, 2019.
\newblock URL \url{https://openreview.net/forum?id=B1VZqjAcYX}.

\bibitem[Li et~al.(2023{\natexlab{a}})Li, Zhang, Koto, Yang, Zhao, Gong, Duan,
  and Baldwin]{li2023cmmlu}
Li, H., Zhang, Y., Koto, F., Yang, Y., Zhao, H., Gong, Y., Duan, N., and
  Baldwin, T.
\newblock Cmmlu: Measuring massive multitask language understanding in chinese,
  2023{\natexlab{a}}.

\bibitem[Li et~al.(2023{\natexlab{b}})Li, Bubeck, Eldan, Del~Giorno, Gunasekar,
  and Lee]{textbooks2}
Li, Y., Bubeck, S., Eldan, R., Del~Giorno, A., Gunasekar, S., and Lee, Y.~T.
\newblock Textbooks are all you need ii: \textbf{phi-1.5} technical report.
\newblock \emph{arXiv preprint arXiv:2309.05463}, 2023{\natexlab{b}}.

\bibitem[Loshchilov \& Hutter(2016)Loshchilov and Hutter]{loshchilov2016sgdr}
Loshchilov, I. and Hutter, F.
\newblock Sgdr: Stochastic gradient descent with warm restarts.
\newblock \emph{arXiv preprint arXiv:1608.03983}, 2016.

\bibitem[Loshchilov \& Hutter(2017)Loshchilov and
  Hutter]{loshchilov2017decoupled}
Loshchilov, I. and Hutter, F.
\newblock Decoupled weight decay regularization.
\newblock \emph{arXiv preprint arXiv:1711.05101}, 2017.

\bibitem[Ma et~al.(2023)Ma, Fang, and Wang]{ma2023llm}
Ma, X., Fang, G., and Wang, X.
\newblock Llm-pruner: On the structural pruning of large language models.
\newblock \emph{arXiv preprint arXiv:2305.11627}, 2023.

\bibitem[Narayan et~al.(2018)Narayan, Cohen, and Lapata]{narayan2018dont}
Narayan, S., Cohen, S.~B., and Lapata, M.
\newblock Don't give me the details, just the summary! topic-aware
  convolutional neural networks for extreme summarization, 2018.

\bibitem[Peiyuan~Zhang \& Lu(2023)Peiyuan~Zhang and Lu]{tinyllama}
Peiyuan~Zhang, Guangtao~Zeng, T.~W. and Lu, W.
\newblock Tinyllama, Sep 2023.
\newblock URL \url{https://github.com/jzhang38/TinyLlama}.

\bibitem[Peng et~al.(2023)Peng, Alcaide, Anthony, Albalak, Arcadinho, Cao,
  Cheng, Chung, Grella, GV, et~al.]{peng2023rwkv}
Peng, B., Alcaide, E., Anthony, Q., Albalak, A., Arcadinho, S., Cao, H., Cheng,
  X., Chung, M., Grella, M., GV, K.~K., et~al.
\newblock Rwkv: Reinventing rnns for the transformer era.
\newblock \emph{arXiv preprint arXiv:2305.13048}, 2023.

\bibitem[Radford et~al.(2018)Radford, Narasimhan, Salimans, Sutskever,
  et~al.]{radford2018improving}
Radford, A., Narasimhan, K., Salimans, T., Sutskever, I., et~al.
\newblock Improving language understanding by generative pre-training.
\newblock 2018.

\bibitem[Radford et~al.(2019)Radford, Wu, Child, Luan, Amodei, Sutskever,
  et~al.]{radford2019language}
Radford, A., Wu, J., Child, R., Luan, D., Amodei, D., Sutskever, I., et~al.
\newblock Language models are unsupervised multitask learners.
\newblock \emph{OpenAI blog}, 1\penalty0 (8):\penalty0 9, 2019.

\bibitem[Ren et~al.(2023)Ren, Zhou, Meng, Huang, Wang, Wang, Li, Zhang,
  Podolskiy, Arshinov, et~al.]{ren2023PanGu}
Ren, X., Zhou, P., Meng, X., Huang, X., Wang, Y., Wang, W., Li, P., Zhang, X.,
  Podolskiy, A., Arshinov, G., et~al.
\newblock Pangu-{$\Sigma$}: Towards trillion parameter language model with
  sparse heterogeneous computing.
\newblock \emph{arXiv preprint arXiv:2303.10845}, 2023.

\bibitem[Shazeer(2019)]{shazeer2019fast}
Shazeer, N.
\newblock Fast transformer decoding: One write-head is all you need.
\newblock \emph{arXiv preprint arXiv:1911.02150}, 2019.

\bibitem[Shibata et~al.(1999)Shibata, Kida, Fukamachi, Takeda, Shinohara,
  Shinohara, and Arikawa]{shibata1999byte}
Shibata, Y., Kida, T., Fukamachi, S., Takeda, M., Shinohara, A., Shinohara, T.,
  and Arikawa, S.
\newblock Byte pair encoding: A text compression scheme that accelerates
  pattern matching.
\newblock 1999.

\bibitem[Sun et~al.(2020)Sun, Yu, Yu, and Cardie]{sun2019investigating}
Sun, K., Yu, D., Yu, D., and Cardie, C.
\newblock Investigating prior knowledge for challenging chinese machine reading
  comprehension.
\newblock \emph{Transactions of the Association for Computational Linguistics},
  2020.
\newblock URL \url{https://arxiv.org/abs/1904.09679v3}.

\bibitem[Tanaka et~al.(2020)Tanaka, Kunin, Yamins, and Ganguli]{prune_synflow}
Tanaka, H., Kunin, D., Yamins, D. L.~K., and Ganguli, S.
\newblock Pruning neural networks without any data by iteratively conserving
  synaptic flow.
\newblock In Larochelle, H., Ranzato, M., Hadsell, R., Balcan, M., and Lin, H.
  (eds.), \emph{Advances in Neural Information Processing Systems 33: Annual
  Conference on Neural Information Processing Systems 2020, NeurIPS 2020,
  December 6-12, 2020, virtual}, 2020.

\bibitem[Tang et~al.(2020)Tang, Wang, Xu, Tao, Xu, Xu, and Xu]{tang2020scop}
Tang, Y., Wang, Y., Xu, Y., Tao, D., Xu, C., Xu, C., and Xu, C.
\newblock Scop: Scientific control for reliable neural network pruning.
\newblock \emph{Advances in Neural Information Processing Systems},
  33:\penalty0 10936--10947, 2020.

\bibitem[Tang et~al.(2024)Tang, Wang, Guo, Tu, Han, Hu, and
  Tao]{tang2024survey}
Tang, Y., Wang, Y., Guo, J., Tu, Z., Han, K., Hu, H., and Tao, D.
\newblock A survey on transformer compression.
\newblock \emph{arXiv preprint arXiv:2402.05964}, 2024.

\bibitem[Team(2023)]{team2023internlm}
Team, I.
\newblock Internlm: A multilingual language model with progressively enhanced
  capabilities, 2023.

\bibitem[Toneva et~al.(2018)Toneva, Sordoni, Combes, Trischler, Bengio, and
  Gordon]{toneva2018empirical}
Toneva, M., Sordoni, A., Combes, R. T.~d., Trischler, A., Bengio, Y., and
  Gordon, G.~J.
\newblock An empirical study of example forgetting during deep neural network
  learning.
\newblock \emph{arXiv preprint arXiv:1812.05159}, 2018.

\bibitem[Touvron et~al.(2023)Touvron, Lavril, Izacard, Martinet, Lachaux,
  Lacroix, Rozi{\`e}re, Goyal, Hambro, Azhar, et~al.]{touvron2023llama}
Touvron, H., Lavril, T., Izacard, G., Martinet, X., Lachaux, M.-A., Lacroix,
  T., Rozi{\`e}re, B., Goyal, N., Hambro, E., Azhar, F., et~al.
\newblock Llama: Open and efficient foundation language models.
\newblock \emph{arXiv preprint arXiv:2302.13971}, 2023.

\bibitem[Vaswani et~al.(2017)Vaswani, Shazeer, Parmar, Uszkoreit, Jones, Gomez,
  Kaiser, and Polosukhin]{vaswani2017attention}
Vaswani, A., Shazeer, N., Parmar, N., Uszkoreit, J., Jones, L., Gomez, A.~N.,
  Kaiser, {\L}., and Polosukhin, I.
\newblock Attention is all you need.
\newblock \emph{Advances in neural information processing systems}, 30, 2017.

\bibitem[Wang et~al.(2020)Wang, Pruksachatkun, Nangia, Singh, Michael, Hill,
  Levy, and Bowman]{wang2020superglue}
Wang, A., Pruksachatkun, Y., Nangia, N., Singh, A., Michael, J., Hill, F.,
  Levy, O., and Bowman, S.~R.
\newblock Superglue: A stickier benchmark for general-purpose language
  understanding systems, 2020.

\bibitem[Wang et~al.(2023)Wang, Chen, Tang, Guo, Han, Nie, Wang, Hu, Bai, Wang,
  et~al.]{wang2023PanGu}
Wang, Y., Chen, H., Tang, Y., Guo, T., Han, K., Nie, Y., Wang, X., Hu, H., Bai,
  Z., Wang, Y., et~al.
\newblock Pangu-$\pi$: Enhancing language model architectures via nonlinearity
  compensation.
\newblock \emph{arXiv preprint arXiv:2312.17276}, 2023.

\bibitem[Wei et~al.(2023)Wei, Zhao, Zhang, Zhu, Wang, Yang, Li, Cheng, Lü, Hu,
  Li, Yang, Luo, Wu, Liu, Cheng, Cheng, Zhang, Zhang, Lin, Wang, Ma, Dong, Sun,
  Chen, Peng, Liang, Yan, Fang, and Zhou]{wei2023skywork}
Wei, T., Zhao, L., Zhang, L., Zhu, B., Wang, L., Yang, H., Li, B., Cheng, C.,
  Lü, W., Hu, R., Li, C., Yang, L., Luo, X., Wu, X., Liu, L., Cheng, W.,
  Cheng, P., Zhang, J., Zhang, X., Lin, L., Wang, X., Ma, Y., Dong, C., Sun,
  Y., Chen, Y., Peng, Y., Liang, X., Yan, S., Fang, H., and Zhou, Y.
\newblock Skywork: A more open bilingual foundation model, 2023.

\bibitem[Winata et~al.(2023)Winata, Xie, Radhakrishnan, Wu, Jin, Cheng,
  Kulkarni, and Preotiuc-Pietro]{winata2023overcoming}
Winata, G.~I., Xie, L., Radhakrishnan, K., Wu, S., Jin, X., Cheng, P.,
  Kulkarni, M., and Preotiuc-Pietro, D.
\newblock Overcoming catastrophic forgetting in massively multilingual
  continual learning.
\newblock \emph{arXiv preprint arXiv:2305.16252}, 2023.

\bibitem[Xia et~al.(2023)Xia, Gao, Zeng, and Chen]{xia2023sheared}
Xia, M., Gao, T., Zeng, Z., and Chen, D.
\newblock Sheared llama: Accelerating language model pre-training via
  structured pruning.
\newblock 2023.

\bibitem[Xu et~al.(2021)Xu, Lu, Yuan, Zhang, Xu, Yuan, Wei, Pan, Tian, Qin,
  et~al.]{xu2021fewclue}
Xu, L., Lu, X., Yuan, C., Zhang, X., Xu, H., Yuan, H., Wei, G., Pan, X., Tian,
  X., Qin, L., et~al.
\newblock Fewclue: A chinese few-shot learning evaluation benchmark.
\newblock \emph{arXiv preprint arXiv:2107.07498}, 2021.

\bibitem[Yang et~al.(2023)Yang, Xiao, Wang, Zhang, Bian, Yin, Lv, Pan, Wang,
  Yan, et~al.]{yang2023baichuan}
Yang, A., Xiao, B., Wang, B., Zhang, B., Bian, C., Yin, C., Lv, C., Pan, D.,
  Wang, D., Yan, D., et~al.
\newblock Baichuan 2: Open large-scale language models.
\newblock \emph{arXiv preprint arXiv:2309.10305}, 2023.

\bibitem[Yi(2023)]{Yi}
Yi.
\newblock A series of large language models trained from scratch by developers
  at 01-ai.
\newblock \url{https://github.com/01-ai/Yi}, 2023.

\bibitem[You et~al.(2017)You, Gitman, and Ginsburg]{you2017large}
You, Y., Gitman, I., and Ginsburg, B.
\newblock Large batch training of convolutional networks.
\newblock \emph{arXiv preprint arXiv:1708.03888}, 2017.

\bibitem[You et~al.(2019)You, Li, Reddi, Hseu, Kumar, Bhojanapalli, Song,
  Demmel, Keutzer, and Hsieh]{you2019large}
You, Y., Li, J., Reddi, S., Hseu, J., Kumar, S., Bhojanapalli, S., Song, X.,
  Demmel, J., Keutzer, K., and Hsieh, C.-J.
\newblock Large batch optimization for deep learning: Training bert in 76
  minutes.
\newblock \emph{arXiv preprint arXiv:1904.00962}, 2019.

\bibitem[Zellers et~al.(2019)Zellers, Holtzman, Bisk, Farhadi, and
  Choi]{zellers2019hellaswag}
Zellers, R., Holtzman, A., Bisk, Y., Farhadi, A., and Choi, Y.
\newblock Hellaswag: Can a machine really finish your sentence?, 2019.

\bibitem[Zeng et~al.(2022)Zeng, Liu, Du, Wang, Lai, Ding, Yang, Xu, Zheng, Xia,
  et~al.]{zeng2022glm}
Zeng, A., Liu, X., Du, Z., Wang, Z., Lai, H., Ding, M., Yang, Z., Xu, Y.,
  Zheng, W., Xia, X., et~al.
\newblock Glm-130b: An open bilingual pre-trained model.
\newblock \emph{arXiv preprint arXiv:2210.02414}, 2022.

\bibitem[Zhong et~al.(2023)Zhong, Cui, Guo, Liang, Lu, Wang, Saied, Chen, and
  Duan]{zhong2023agieval}
Zhong, W., Cui, R., Guo, Y., Liang, Y., Lu, S., Wang, Y., Saied, A., Chen, W.,
  and Duan, N.
\newblock Agieval: A human-centric benchmark for evaluating foundation models,
  2023.

\end{thebibliography}
\bibliographystyle{icml2024}

\newpage
\appendix
\onecolumn

\section{Multi-head Attention v.s. Group-query Attention}
\label{convert}

Current mainstream LLaMA architectures, such as Qwen~\cite{qwen} and InternLM~\cite{team2023internlm}, commonly employ multi-head attention (MHA) mechanisms~\cite{vaswani2017attention}. However, storing a large amount of KV (key and value) cache in memory-constrained edge devices poses a significant challenge, especially in long text input scenarios. Group query attention~\cite{ainslie2023gqa} is recently proposed to seek a balance between multi-head attention and multi-query attention~\cite{shazeer2019fast}. By sharing the same key-value heads for all queries in each group, the RAM requirement for the KV cache can be largely reduced in edge devices. We convert our pretrained PanGu-$\pi$-1.5B Pro to its GQA version by mean-pooling~\cite{ainslie2023gqa} the KV heads in each group and then continually training on only 5\% of the origin dataset. The number of groups is set to 8. As shown in Table~\ref{tab:gqa}, the converted GQA version model exhibits comparable performance to the MHA counterpart with fewer parameters.

\begin{table}[ht]
	\centering
%	\vspace{-5pt}
	\caption{Comparison between MHA and GQA.}
	\label{tab:gqa}%
	\begin{tabular}{cc|cccc}
		\toprule
		Attn. & Size & C-Eval & CMMLU & MMLU & AGI-Eval     \\
		\midrule
		MHA     & 1.5B   & 52.91 & 49.51 & 53.76 & 44.42   \\    
		GQA     & 1.4B   & 48.75 & 46.94 & 51.97  & 43.59\\
		\bottomrule
	\end{tabular}%
	%    \vspace{-10pt}
\end{table}%

\section{Improved Random Initialization}
\label{sec:a-random}

\begin{figure*}[b]
	%\vskip 0.2in
	\begin{center}
		\centerline{\includegraphics[width=0.99\linewidth]{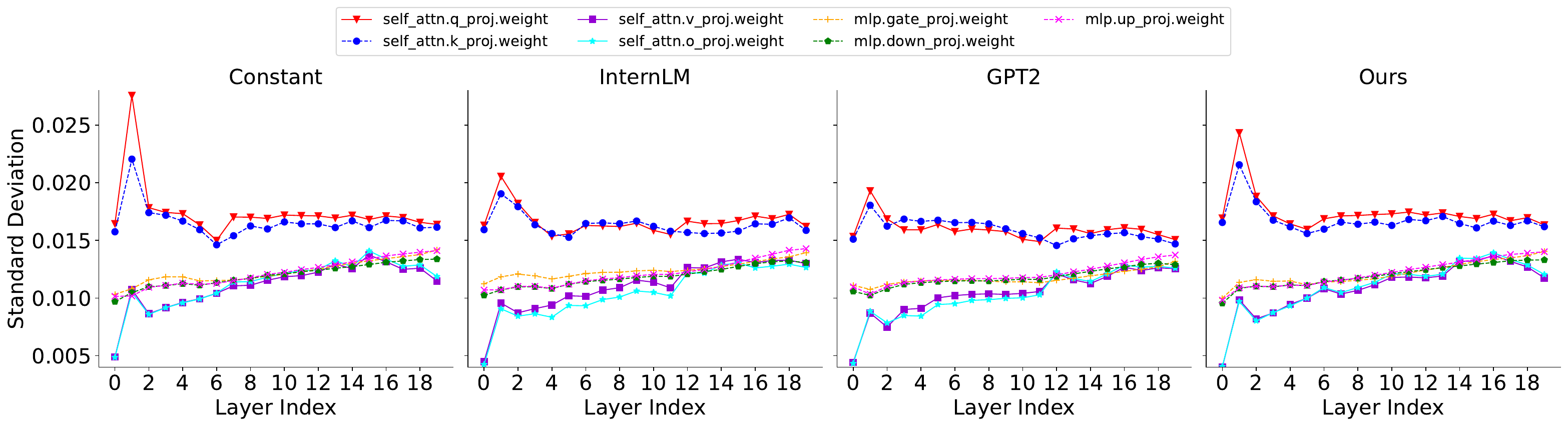}}
%		\vspace{-12pt}
		\caption{The standard deviations of different layers after pretraining.}
		\label{img:init_ours}
	\end{center}
	%	\vskip -0.2in
\end{figure*}

When training the model from scratch, we typically initialise the weight of linear layers using a normal distribution $N(0,\sigma^2)$ with zero mean and standard deviation $\sigma$.
Some methods~\cite{radford2019language, team2023internlm} use different standard deviations for different layers.
Figure~\ref{img:init_ours} shows the standard deviations of different layers after pretraining, indicating that
the parameter distributions will be similar from different initialisation values.
In particular, the variance of the four linear layers within the MHA layer varies with the depth, while the variance of the linear layers within the MLP layers remains almost constant.

We introduce an improved initialization method that adapts the standard deviation of different linear weights to the depth of the model. In the MHA, the initial standard deviation of the query and key projection parameters decreases from $\sqrt{2}\sigma$ to $\sigma$, while that of the value and out projection parameters increases from $\sigma/\sqrt{2}$ to $\sigma$, as the layer index increases. In the MLP, we keep the initial standard deviation of the linear weights constant at $\sigma$ for all layers. As demonstrated in Table~\ref{tab:init_pref_2}, our approach exhibits a marginal superiority over the constant initialization approach.

\begin{table}[t]
	\centering
	\caption{Performance under different initialization strategies. Our method exhibits a slight edge over the constant standard deviation approach.}
	%\label{tab:init_pref}%
	\begin{tabular}{l|ccc|c}
		\toprule
		Initialization Method  & ARC-E & HellaSwag & C3 & Avg.   \\
		\midrule
		Constant & 37.57  & 41.16  & 49.04  & 42.59  \\ 
		%        Xavier~\cite{} & 34.39  & 40.10  & 47.70  & 40.73  \\ 
		GPT2~\cite{radford2019language} & \textbf{38.62} & 39.34  & 48.44  & 42.13  \\ 
		InternLM~\cite{team2023internlm} & 34.39 & 41.48  & 47.70 & 41.19 \\ 
		Ours & 37.57  & \textbf{42.00}  & \textbf{49.26}  & \textbf{42.94} \\ 
		\bottomrule
	\end{tabular}
	\label{tab:init_pref_2}
	%   \vspace{-12pt}
\end{table}

\section{The Impact of Attention Head}

Table~\ref{tab:attnetion_head_num} shows how performance varies \wrt~the number of attention heads. The results show that it  does not affect the inference speed or the model performance significantly, as long as the hidden size is fixed. The head dimension is set to 128 in PanGu-$\pi$-1B Pro and PanGu-$\pi$-1.5B Pro.

\begin{table}[h]
	\centering
	\caption{Varying the number of attention heads.}
	\label{tab: head_dim}%
	\begin{tabular}{ccc|ccc|c}
		\toprule
		Heads & Head Dimension & Speed & ARC-E & HellaSwag   & C3  & Avg.  \\
		\midrule
		14     & 128  &  29.49  & 34.39 & 41.48 & 47.70 & 41.19 \\
		28     & 64  &    30.11   & 35.39  & 41.63  & 48.09  & 41.70  \\
		56      & 32   &  30.49  & 33.16  & 41.36  & 48.17  & 40.90 \\
		\bottomrule
	\end{tabular}%
	\label{tab:attnetion_head_num}
\end{table}%

\section{Weight Decay}

Weight decay~\cite{loshchilov2017decoupled} is a commonly employed regularization method aimed at mitigating overfitting on the training set. We delve into its impact in Table~\ref{tab:weight_decay}. Elevating the weight decay imparts more robust regularization, albeit at the expense of constraining the model's representation capacity. Through empirical experiments, we observe that the model attains optimal performance when the weight decay is set at 0.1.

\begin{table}[h]
	\centering
	\caption{Performance under different weight decay. The model achieved the best performance with a weight decay of 0.1.}
	\begin{tabular}{c|ccc|c}
		\toprule
		Weight Decay & ARC-E & HellaSwag  & C3& Average \\
		\midrule
		0.2 & 34.68 & 36.15 & 45.31 & 38.71 \\
		0.1 & 34.39 & \textbf{41.48}  & \textbf{47.70} & \textbf{41.19} \\ 
		0.01 & \textbf{34.74} & 36.76 & 45.26 & 38.92  \\ 
		0.001 & 33.59 & 37.07 & 44.93 & 38.53  \\ 
		0.0001 & 31.22 & 37.76 & 44.11 & 37.70 \\ 
		\bottomrule
	\end{tabular}%
	\label{tab:weight_decay}%
\end{table}%

\newpage

\section{Additional Results of Layer Selection}
\begin{figure*}[ht] 
	%\vskip 0.2in
	\begin{center}
		\centerline{\includegraphics[width=0.99\textwidth]{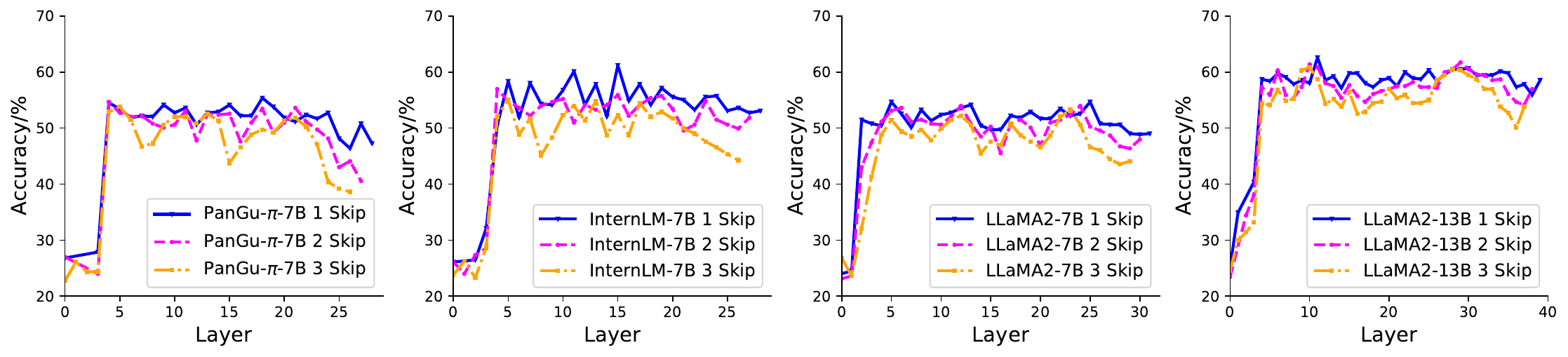}} 
		\centerline{\includegraphics[width=0.99\textwidth]{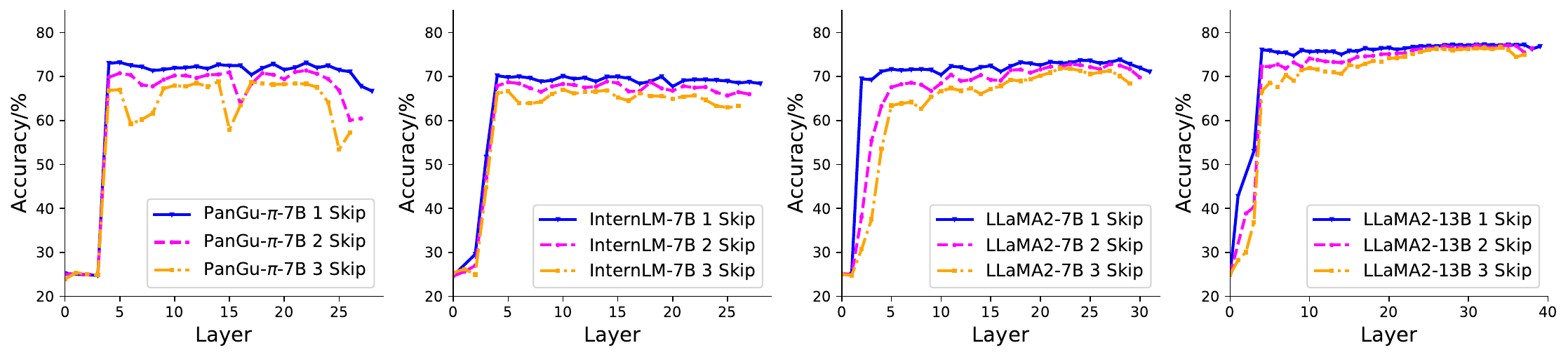}} 
		\centerline{\includegraphics[width=0.99\textwidth]{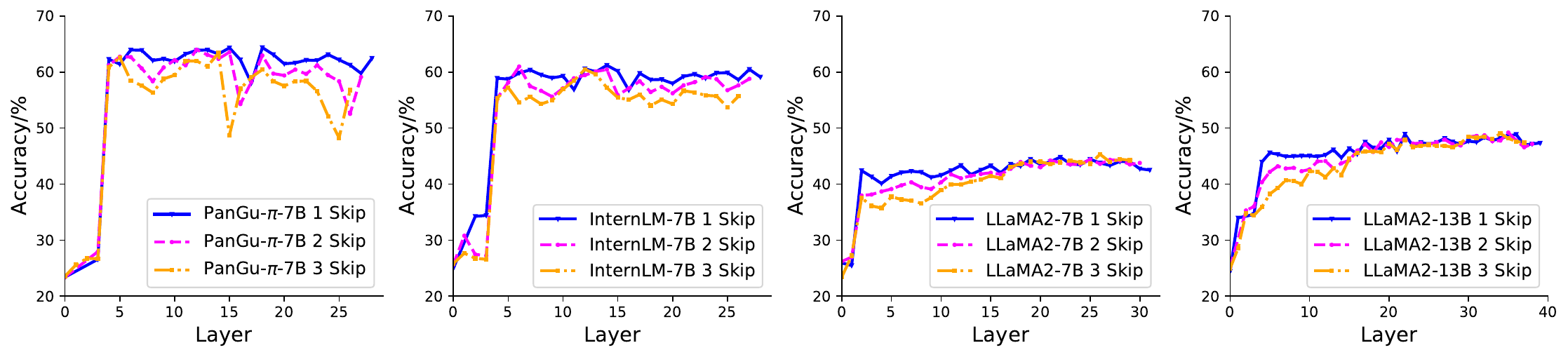}} 
%		\vspace{-15pt}
		\caption{Layer-skipped performance of large language models on single downstream tasks. 
			From top to bottom, the tasks are  ARC-E, HellaSwag, and C3, respectively.}
		\label{img:layer_skip2}
	\end{center}
	%	\vskip -0.2in
\end{figure*}

The layer skipping results of single downstream tasks are released in Figure~\ref{img:layer_skip2}. The performance trend of single downstream tasks are consistent to the average results shown Figure 5 of the main paper. Layers situated near the beginning and end of the model often carry more significance than the intermediate layers. 

\end{document}